\documentclass[sw]{iosart2x}
\pubyear{2023}
\volume{0}
\firstpage{1}
\lastpage{43}

\usepackage{graphicx}
\usepackage{multirow}
\usepackage{multicol}
\usepackage[table]{xcolor}
\usepackage{tikz}
\usepackage{todonotes}
\usepackage{tabularx}
\usetikzlibrary{positioning,shapes}
\usepackage{hyperref}
\usepackage{cleveref}
\usepackage{svg}
\usepackage{adjustbox}
\usepackage{lscape}

\definecolor{Gray}{gray}{0.9}

\usepackage{listings}
\usepackage{svg}
\usepackage[euler]{textgreek}
\usepackage{wasysym}
\usepackage{amssymb}
\usepackage[utf8]{inputenc}
\usepackage{xspace}
\usepackage{pifont}
\usepackage{caption}
\usepackage{lipsum}

\DeclareUnicodeCharacter{25CF}{\simple}

\newcommand\circledmark[1][]{%
  \ooalign{%
    \hidewidth
    \kern0.65ex\raisebox{-0.9ex}{\scalebox{3}{\textcolor{red}{\textbullet}}}
    \hidewidth\cr
    $\checkmark$\cr
  }%
}

\begin{document}

\begin{frontmatter}

\title{Construction of Knowledge Graphs: State and Challenges}
\runtitle{Construction of Knowledge Graphs: State and Challenges}

\begin{aug}
\author[B]{\inits{M.}\fnms{Marvin} \snm{Hofer}
\ead[label=e1]{hofer@informatik.uni-leipzig.de}
\thanks{Corresponding author. \printead{e1}.}}
\author[B]{\inits{D.}\fnms{Daniel} \snm{Obraczka}
\ead[label=e2]{obraczka@informatik.uni-leipzig.de}}
\author[A,B]{\inits{A.}\fnms{Alieh} \snm{Saeedi}\ead[label=e3]{saeedi@informatik.uni-leipzig.de}}
\author[C]{\inits{H.}\fnms{Hanna} \snm{Köpcke}\ead[label=e4]{koepcke@informatik.uni-leipzig.de}}
\author[A,B]{\inits{E.}\fnms{Erhard} \snm{Rahm}\ead[label=e5]{rahm@informatik.uni-leipzig.de}}
\address[A]{\orgname{Dept. of Computer Science, Leipzig University}, 
\cny{Germany}}
\address[B]{\orgname{Center  for Scalable Data Analytics and
Artificial Intelligence (ScaDS.AI) Dresden/Leipzig}, 
\cny{Germany}}
\address[C]{\orgname{
Faculty Applied Computer Sciences and Biosciences, Mittweida University of Applied Sciences}, 
\cny{Germany}}
\end{aug}

\begin{abstract}
With knowledge graphs (KGs) at the center of numerous applications such as recommender systems and question answering, the need for generalized pipelines to construct and continuously update such KGs is increasing. While the individual steps that are necessary to create KGs from unstructured (e.g. text) and structured data sources (e.g. databases) are mostly well-researched for their one-shot execution, their adoption for incremental KG updates and the interplay of the individual steps have hardly been investigated in a systematic manner so far. In this work, we first discuss the main graph models for KGs and introduce the major requirements for future KG construction pipelines. Next, we provide an overview of the necessary steps to build high-quality KGs, including cross-cutting topics such as metadata management, ontology development, and quality assurance. We then evaluate the state of the art of KG construction w.r.t the introduced requirements for specific popular KGs as well as some recent tools and strategies for KG construction. Finally, we identify areas in need of further research and improvement. 
\end{abstract}


\end{frontmatter}

\section{Introduction}
\label{sec:introduction}

Aggregated machine-readable information in the form of knowledge graphs (KG) serves as the backbone of numerous data science applications nowadays, ranging from question-answering~\cite{DBLP:conf/wsdm/HuangZLL19} over recommendation systems~\cite{Wang2019KGATKG} to predicting drug-target interactions~\cite{10.1093/bioinformatics/btz600}.
The ever-changing nature of information necessitates the design of KG construction pipelines that are able to incorporate new information continuously.
In the development of such a system, knowledge engineering teams have to deal with a variety of challenges from tackling scalability and heterogeneous data sources to tracking the provenance of data.
Given the usually large volume of data that needs to be integrated, such pipelines have to be automatized as much as possible while aiming at a high degree of data quality.

Knowledge graphs generally integrate heterogeneous data from a variety of sources with unstructured and semi-structured data of different modalities (e.g., pictures, audio, text) as well as structured data such as databases or other KGs in a semantically rich way.
The construction of a KG therefore encompasses a multi-disciplinary effort requiring expertise from research areas such as natural language processing (NLP), data integration, knowledge representation, and knowledge management.
Knowledge graphs are at the center of numerous use-cases for data analysis and decision support. In the clinical setting, enriching patient data with medical background knowledge enables improved clinical decision support~\cite{MedicalBackgroundKnowledge}. Sonntag et. al.~\cite{ClinicalDataIntelligence} argue that properly aligning the semantic labels attached to the patient data with medical ontologies is crucial in creating meaningful access to the heterogeneous patient data.
Knowledge graphs are also used to organize the relevant information for fast emerging global topics, such as pandemics (e.g., Covid-19) or natural disasters~\cite{FanKGCGeoHazards}.
Machine learning also benefits from KGs as a source of labeled training data or other input data~\cite{nickel2015review,hogan2021knowledge} thereby supporting the development of knowledge- and data-driven AI approaches~\cite{ji2021survey}. 
KGs can further be combined with Large Language Models (LLMs)  to improve  factual correctness and explanations in question-answering, e.g. with ChatGPT, thereby promoting quality and interpretability of AI decision-making~\cite{Pan2023UnifyingLL,Yang2023ChatGPTIN}.

Improving the pipelines that build such knowledge graphs and enabling them to efficiently keep current and semantically meaningful aggregated knowledge is therefore an effort that benefits a wide range of application areas. However, KGs are currently created often in a batch-like manner so that the respective pipelines are unfit to incorporate new incoming facts into a KG without full re-computation of the individual tasks. Furthermore, different steps of the pipelines often require manual intervention, thereby limiting scalability to large data volumes and increasing the time for updating a KG.

There is a growing number of surveys about knowledge graphs, especially on their general characteristics and usage forms~\cite{nickel2015review,ji2021survey,hogan2021knowledge}.  
An excellent tutorial-style overview about the construction and curation of KGs is provided in~\cite{Weikum2021MachineKC} with a focus on integrating data from textual and semi-structured data sources such as Wikipedia. 
Other surveys focus on KG construction with 
specific technologies~\cite{Zhu2022MultiModalKG,ryen2022building} or only a single domain such as for geographical data~\cite{Ma2021KnowledgeGC}.
We discuss  related surveys on KG construction in Section~\ref{sec:rel-work} and contrast them with our approach.

This survey article provides a concise yet comprehensive entry into the current state of the art in KG construction for readers new to the topic, as well as contributing valuable guidance for researchers, engineers, and experts by highlighting existing solution approaches, tools, and identifying open gaps in the areas.
We first outline the main requirements for the construction and continuous maintenance of KGs distilled from the literature as well as our experience and reasoning. Next, we give an overview of the concrete subtasks of KG construction and current solution approaches. Furthermore, we select 23 KG-specific construction approaches as well as generic toolsets based on the criteria discussed in Section~\ref{sec:kg-overview} and evaluate and compare them w.r.t. the requirements introduced in Section~\ref{sec:requirements}. Finally, we identify open challenges and current limitations and thus areas for further research.

Our survey builds on  previous studies for KG construction but differs in essential aspects as will be explained in detail in  Section~\ref{sec:rel-work}. In contrast with most other surveys, we explicitly specify the main requirements for KG construction and use these  as a guideline for evaluating current solutions and identifying open challenges. We are also more comprehensive in several important aspects as we cover different graph data models (RDF and property graphs) and deal with incremental KG construction and data integration including incremental entity resolution in much more detail. We also provide a comparison between many carefully selected KG-specific construction approaches and toolsets and identify open challenges that go beyond those discussed in previous surveys. 

The further structure of this survey is as follows: 
\begin{itemize}
   \item After a definition of KGs and a comparative discussion of graph data models for KGs we introduce and categorize the general requirements for incremental KG construction in Section~\ref{sec:preliminaries}.
    \item In Section~\ref{sec:tasks} we provide an overview of the main tasks in incremental KG construction pipelines and proposed solution approaches for them.
    \item We then investigate and compare existing construction efforts for selected KGs as well as within recent tools for KG construction w.r.t. the requirements introduced earlier. This also allows us to identify tasks that are not yet supported well. 
    \item Section~\ref{sec:challenges} discusses open challenges for KG construction. 
    \item Section~\ref{sec:rel-work} contains a more in-depth comparison with the related work.
     \item Finally, we offer concluding remarks and a summary.
\end{itemize}

\section{KG background and requirements for KG construction}
\label{sec:preliminaries}

We first outline the notion of knowledge graph (KG) used in this paper which is based on the integration of information from multiple sources. We then briefly introduce and compare the two most popular graph data models for KGs, namely RDF and property graphs. Finally, we outline the main requirements or desiderata for largely automatic construction and maintenance of KGs.

\subsection{Knowledge Graph}
KGs typically realize physical data integration, where the information from different sources is combined in a new graph-like representation\footnote{While we will focus on the predominant physical data integration of KGs, there are also some virtual data integration approaches, e.g., to keep data source more autonomous~\cite{Xiao2019VirtualKG,DeclarativeAsscheReview23}.}.
KGs are schema-flexible and can thus easily accommodate and interlink heterogeneously structured entities. 
This is in contrast to the use of data warehouses as a popular approach for physical data integration. Data warehouses focus on integrating data within a structured (relational) database with a relatively static schema optimized for certain multi-dimensional data analysis. 
Schema evolution is a manual and tedious process making it difficult to add new data sources or new kinds of information not conforming to the schema.  
KGs are less restricted and can better deal with heterogeneous information derived from semi- and unstructured data from potentially many sources. 

Although the term \textit{knowledge graph} goes back as far as 1973~\cite{schneider1973course}, it gained popularity through the 2012 blog post\footnote{\url{https://blog.google/products/search/introducing-knowledge-graph-things-not/}} 
about the Google KG.
Afterward, several related definitions of knowledge graphs were proposed, either in research papers~\cite{paulheim2017knowledge,ehrlinger2016towards,hogan2021knowledge,Weikum2021MachineKC,lissandriniknowledge} or by companies using or supporting KGs (OpenLink, Ontotext, Neo4J, TopQuadrant, Amazon, Diffbot\footnote{\url{https://blog.diffbot.com/knowledge-graph-glossary/}}, Google). 
Ehrlinger et al.~\cite{ehrlinger2016towards} give a comprehensive overview of KG definitions and provide their own: 
"A knowledge graph acquires and integrates information into an ontology and applies a reasoner to derive new knowledge."
Hogan et al.~\cite{RedefiningKGs} argue that this definition is very specific and excludes various industrial KGs which helped to popularize the concept. 

We therefore define KGs more inclusively as a graph of data consisting of semantically described entities and relations of different types that are integrated from different sources. 
Entities have unique identifier. 
KG entities and relations can be semantically described by an ontology~\cite{feilmayr2016analysis}.
A KG's ontology defines the concepts, relationships, and rules governing the \textit{semantic structure} within a KG of one or several domains that also include the types and properties of entities and their relationships. 
To structure data in a KG, common ontology relationships euch as \textit{is-a} and \textit{has-a} are used to represent taxonomic hierarchies and possessive relations between entities. 
Furthermore, an ontology can enable the inference of new implicit knowledge from the explicitly represented information in the KG~\cite{lissandriniknowledge}.

\begin{figure}
\centering
\includegraphics[width=\textwidth]{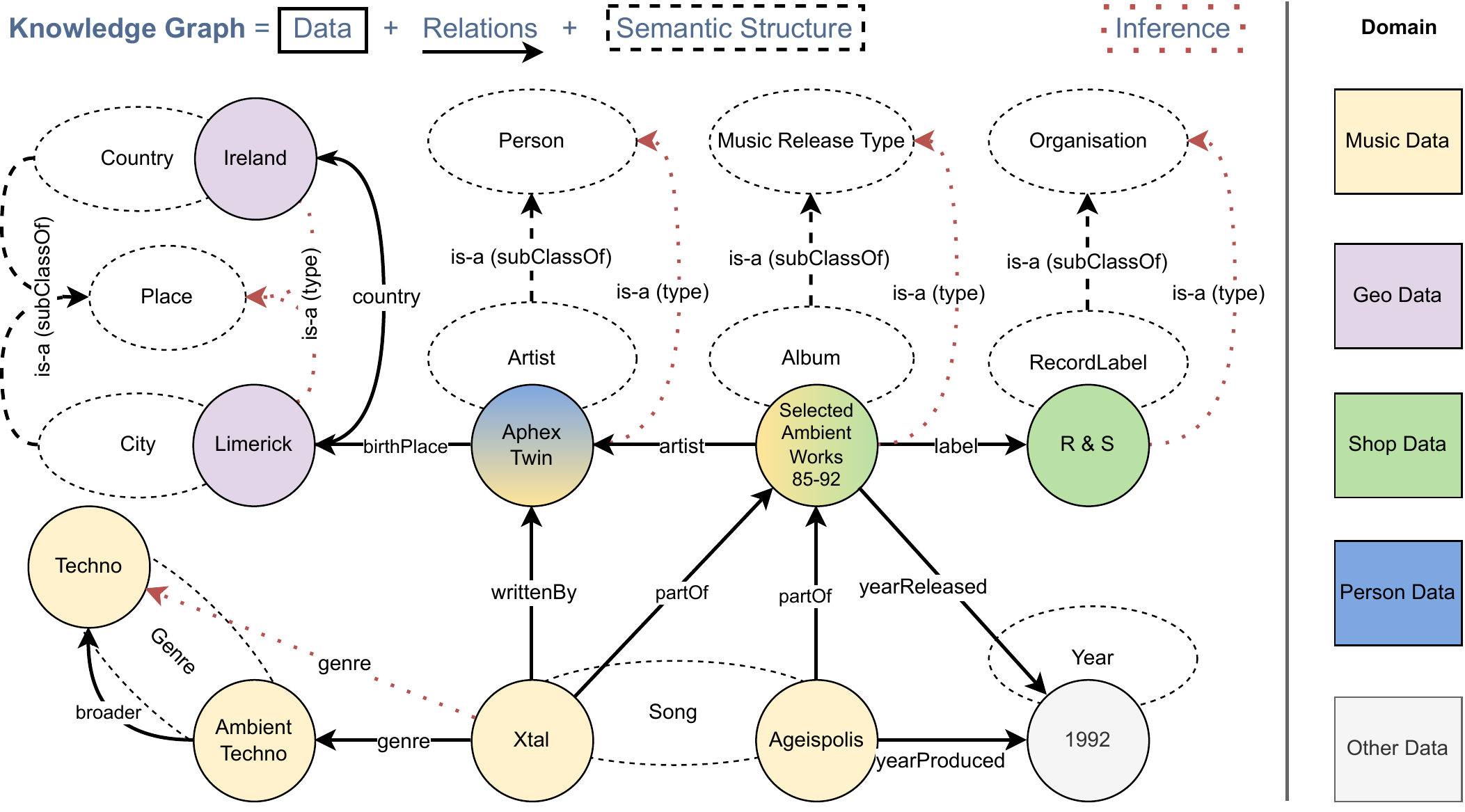}
\caption{Simplified Knowledge Graph (KG) example demonstrating integrated information from five domains, showcasing ten entities of eight types connected by twelve relationships (two distinct is-a relations). Dashed lines indicate semantic structures (ontology or graph schema) such as entity types. Inferences can be made based on the relationships and typing, revealing additional information such as the broader birthplace of Aphex Twin being Ireland and Xtal belonging to the Techno genre (Not all possible inferences are shown for clarity).}
\label{fig:kgdef}
\end{figure}

Figure~\ref{fig:kgdef} visualizes a simplified KG example with integrated information from several domains where ontological information such as types or is-a relations are dashed. 
There are ten entities of the following eight types: Country (\textit{Ireland}), City (\textit{Limerick}), Artist (\textit{Aphex Twin}), Album (\textit{Selected Ambient Works 85-9}), Record Label (\textit{R \& S}), Genre (\textit{Techno}, \textit{Ambient Techno}), Song (\textit{Xtal}, \textit{Ageispolis}) and Year (\textit{1992}).
Ontological \textit{is-a} (sub-class) relations interrelate City and Country with Place, Artist with Person, Album with Music Release Type, and Record Label with Organisation.
The domain is further described by the named relationships: country, birthPlace, artist, label, writtenBy, yearReleased, founded, broader, genre, yearProduced, partOf. 
Based on the given relationships and typing, further information is inferable, e.g., Aphex Twin's broader birthplace is Ireland, the song Xtal is also of genre Techno, Aphex Twin being of the type Artist means this instance is also of the type Person (for readability, not all possible inferences are denoted).

\subsection{Graph Models}
\label{sec:models}
\begin{adjustbox}{center,caption={Comparison of RDF and property graphs.},float=table}
\footnotesize
\label{tab:rdf-pgm}

\begin{tabularx}{1\textwidth}{|l|X|X|}
\hline
& Resource Description Framework (RDF) & Property Graph Model (PGM) \\
\hline
base constructs & triples \texttt{<}subject, predicate, object\texttt{>}
& labeled vertices and edges and their properties \\
entity identity & IRI-based & local (implementation-specific) \\
node classification & rdf:type triples & type labels \\
ontology support & RDFS, OWL2 vocabularies & limited, e.g., schema graph \\
integrity constraints & SHACL, SHEX & PG-Keys, PG-Schema \\
query language & SPARQL(-Star) & Cypher, Gremlin, G-Core, PGQL \\
exchange format & N-Triples, N-Quads, (RDF/XML, JSONLD) & application specific e.g., PGEF, GDL \\
meta information & reification, singleton-property, (RDF-Star) & dedicated properties \\ 
\hline
\end{tabularx}

\end{adjustbox}

To represent and use KGs as informally defined above, a powerful graph data model is needed that supports entities and relations of different types as well as their ontological description and organization~\cite{DBLP:journals/cacm/SakrBVIAAAABBDV21}. Moreover, the graph data model should provide a comprehensive query language and possibly more advanced graph analysis or mining capabilities, e.g., for clustering of similar entities or determining graph embeddings for machine learning tasks. Support for integrity constraints is also desirable to automatically control the consistency and therefore, quality of graph data to some extent. 
Furthermore, it should be possible to represent annotating metadata of KG entities, e.g., about their origin and transformation during KG construction. It is also desirable to reflect the development of the KG over time so that a temporal KG analysis is supported. This can be achieved by a temporal graph data model with time metadata for every entity and relation and temporal query possibilities, e.g., to determine previous states of the KG or to find out what has been changed in a certain time interval. 
The temporal development of a KG might alternatively be reflected with a versioning concept where new KG versions are periodically released. Finally, the graph data model should facilitate the KG construction process and its different tasks for acquiring, transforming and integrating heterogeneous data from different sources. This can be supported by suitable formats to seamlessly exchange data of the chosen graph model between different steps and processing nodes of a KG construction pipeline. 

The most common graph models used for KGs are the Resource Description Framework and the Property Graph Model. 
In the following, we briefly describe both and discuss how they meet the introduced desiderata. 
Table~\ref{tab:rdf-pgm} summarizes some of the key differences of both models. At the end, we also contrast the different terminology of the models and specify the terms used in the rest of this paper. 

\vspace{\baselineskip}
\textbf{Resource Description Framework}
(RDF) is a framework or data model to present data in a graph-like fashion, and was initially developed to describe metadata of web-resources (namely the Semantic Web)~\cite{lassila1999resource}. 
Today, the W3C proposes many technologies around RDF that help build and use knowledge graphs either as part of the Linked Data Cloud or in an encapsulated environment. KGs are represented by a set of <subject, predicate, object> triples that uniformly represent named relations (predicates) of entities (subjects) to either attribute values (literals) or other entities (objects). 
Entities are usually assigned an IRI (Internationalized Resource Identifier) that can refer to either a global or local namespace. In addition to IRIs, entities can identify by a blank node identifier that is only unique inside an RDF dataset.
Sets of triples may also be grouped within named graphs to aggregate more information by extending the triple structure to quads of the form <subject, predicate, object, named-graph>. 
Standard RDF does not support edge properties, although the RDF-Star extension\footnote{\url{https://www.w3.org/2022/08/rdf-star-wg-charter/}} includes a similar feature, where single triples are usable in the subject or object part of another triple. 

The RDF standard also defines vocabularies such as RDF Schema (RDFS) to further express semantic structure by allowing the definition of classes, properties and their hierarchies. 
An RDF resource's type (or entity class) is assigned by using the standard RDF vocabulary\footnote{\url{http://www.w3.org/1999/02/22-rdf-syntax-ns\#}} to define triples of the form \texttt{<s rdf:type o>} where o is the class (type) of the resource.
In addition to RDFS, a widely used approach for defining ontologies is the Web Ontology Language (OWL), more specifically, the current version OWL 2, which adds semantics to the data using a variety of axioms.

Besides syntax validation (triples/quads, URIs, datatypes) RDF triple stores do not provide a standard method to define and validate graph data integrity or shape constraints (similar to relation database schemata). Therefore overlaid solutions such as SHACL (Shape Constraint Language~\cite{shacl}) or ShEx (Shape Expressions~\cite{Prudhommeaux2014ShapeEA}) are developed that can be used to validate the semantic correctness of the graphs structure, node or property constraints, cardinalities, and other constructs.

While some RDF stores or triple stores are built from scratch to optimize the management of RDF triples, others might use existing SQL or NoSQL systems in the underlying database processing layer. 
The primary query language for RDF (moreover, the Semantic Web) is the standardized language SPARQL\footnote{\url{https://www.w3.org/TR/sparql11-overview/}}, with an extended version for RDF-Star called SPARQL-Star.
Standard exchange formats for RDF are N-Triples, N-Quads, Turtle, or adapted syntax formats like RDF/XML and JSON-LD.

There are different possibilities to assign metadata to entities, relations and properties like using RDF-Star or named graphs as explained and evaluated in~\cite{frey2019evaluation, sikos2020provenance}.
The usage of support constructs for metadata management generally increases the complexity of the graph structure and queries and can possibly increase processing time.

There is some work around the representation, querying storage and other aspects of temporal information in RDF~\cite{Zhang2021RDFFT}. The investigated methods focus on different temporal granularity and dimensions, including approaches that target to query single snapshots, time windows or to inspect the evolution of temporal graphs.

As many knowledge graphs are in RDF, several frameworks have been developed to perform graph analytics, algorithms, or mining tasks using RDF as input~\cite{Lehmann2017DistributedSA}.
    
\vspace{\baselineskip}
\textbf{Property Graph Model (PGM)}~\cite{Angles2018ThePG}.
The property graph data model, also called Labeled Property Graph (LPG), supports the flexible definition of graph structures with heterogeneous nodes (vertices) and directed edges to represent entities of different kinds, and the relationships between them. Both nodes and edges can have multiple (type-)labels expressing their role in the modeled domain, e.g., \textit{User} as a node label and \textit{follows} as edge label. Additionally, properties (in the form of key-value pairs) can be assigned to both nodes and edges. Further, in the most common implementations, vertices and edges are specified by a unique identifier. 
While label and property names represent some schema-like information, there is intentionally no predefined schema to allow flexible incorporation of heterogeneous entities and relations of different kinds (although a schema graph can be inferred from the type information~\cite{lbath2021schema}). There is no built-in support for ontologies, e.g., to provide is-a relations between entity categories. 
Embedded metadata can be relatively easy maintained for entities and relationships by using dedicated properties, e.g. for provenance or time annotations. 

In contrast to RDF, the PGM with its vertices, edges and properties is more related to graph models in graph theory thereby contributing to their good understandability. 
As there is not yet a global (defacto) standard for PGM, its capabilities highly depend on its implementation. 
The PGM is increasingly popular in research and practice and supported by several graph database systems, such as Neo4j~\cite{Neo4j}, JanusGraph~\cite{JanusGraph} or TigerGraph~\cite{TigerGraph}, and processing frameworks, such as Oracle Labs PGX~\cite{Hong15PGXd} or Gradoop~\cite{rost2021distributed}. 
It is further the base data model for several graph query languages~\cite{DBLP:journals/sigmod/Wood12}, such as G-Core~\cite{DBLP:conf/sigmod/AnglesABBFGLPPS18}, Gremlin~\cite{gremlin}, PGQL~\cite{DBLP:conf/grades/RestHKMC16}, Cypher~\cite{francis2018cypher}, as well as SQL/PGQ and GQL~\cite{DBLP:conf/sigmod/DeutschFGHLLLMM22}, the upcoming ISO standard language for property graph querying. 
Efforts on a standardized PGM serialization format, are the JSON-based Property Graph Exchange Format (PGEF)~\cite{chiba2019property}, YARS-PG~\cite{DBLP:conf/bdas/TomaszukASLC19} or the Graph Definition Language (GDL)\footnote{\url{https://github.com/dbs-leipzig/gdl}}.
    
Similar to RDF stores, data integrity of PGM databases is generally limited to syntax or basic value constraints. 
A first effort about the aspects of property graph \textit{key constraints} is proposed by Angles et al.~\cite{Angles2021PGKeysKF} by identifying four natural key types: identifier, exclusive mandatory, exclusive singleton, or exclusive. Additionally, PG-Schemas offer a robust formalism for specifying property graph schemas~\cite{Bonifati2022PGSchemaSF}.

There are several extensions to the PGM for supporting temporally evolving graph data~\cite{rost2021distributed,rost2021bitemporal} and graph streams~\cite{DBLP:journals/corr/abs-1912-12740}, often with advanced analysis capabilities for graph mining. 

\vspace{\baselineskip}
\textbf{Discussion.}
As outlined above there are pros and cons for both RDF and PGM. 
The intensive use of RDF in the Semantic Web and Linked Open Data communities has lead to its wide-spread application for KGs; in fact most KGs we will consider in Section~\ref{sec:example-kgs} are using RDF. The triple-based graph representation of RDF is quite flexible and allows a uniform representation for entities and relationships. But it is also hard to understand without additional processing or inference as the information of an entity is distributed over many triples. 
While RDF-Star greatly improves the formal meta expressiveness of RDF, specific cases are still not presentable as in PGM without utilizing support constructs. 
In the PGM, we can have two relations with the same name that can be addressed independently. Each relation has its own distinct properties. However, in RDF-Star, relations (triples) are identified based on their associated elements $<<$s1,p1,o1$>$,p2,o2$>$, and it is not possible to attach different sets of information to equally named relations (triples) without causing incorrect connections or relying on support constructs (e.g. singleton properties)~\cite{lassila2022onegraph}.
While RDF is older and has gone through extensive standardization during the last 25 years, the PGM has become increasingly popular for advanced database and network applications, such as graph traversal and network analysis~\cite{WorldOfGraphDBIndustry2022}.

Besides RDF (direct graphs) or property graphs, in some cases, custom models or special high arity representation could be used to cover specific features, such as access-levels, temporal information, or multihop relations in one record (node-edge-edge-node)~\cite{ilyas2022saga}.
However, the usage of such custom models will lower interoperability with existing tools (requiring transformation), and complicate its own reusability by others.

The decision between the use of RDF and PGM (or a custom data model) depends on the targeted application or use case of the final knowledge graph. 
Lassila et al~\cite{lassila2022onegraph} conclude that both formats are qualified to meet their challenges and none of the two is perfect for every use case. They thus recommend increasing interoperability between both models to reuse existing techniques of both approaches.
Various efforts to address this problem have been made in recent years.
The Amazon Neptune\footnote{\url{https://aws.amazon.com/en/blogs/aws/amazon-neptune-a-fully-managed-graph-database-service/}} database service allows users to operate PGM and RDF interchangeably. Hartig et al.~\cite{ReconciliationRDFPG} and Abuoda et al.~\cite{Abuoda2022TransformingRT} discuss transformation strategies between RDF and PGM to lower usage boundaries. GraphQL\footnote{\url{https://graphql.org/}} provides a unified approach to query both RDF and the PGM, although with fewer  features compared to query languages dedicated to these graph formats. GraphQL-LD~\cite{GraphQLLD} aims at simplifying querying Linked Data via GraphQL.

\vspace{\baselineskip}
\textbf{Terminology.}
Due to the different communities around PGM and RDF there are many similar but differently named terms in use. 
Table~\ref{tab:kg-terms} lists some of the terms that we will use as synonymous in this paper with the underlined ones used preferably. 

\begin{adjustbox}{center,caption={Synonymously used KG terms in RDF and PGM.},float=table}
\footnotesize
    % \caption{Synonymously used KG terms in RDF and PGM.}
    \label{tab:kg-terms}
    % \centering
    \begin{tabular}{|l|l|}
    \hline
    Terms & Description \\
    \hline
    \underline{entity}, instance, subject \& object \& resource (RDF), individual & KG nodes that represent a specific real-world or abstract thing \\ % Node, Vertex
    \underline{relation}, property (RDF) & a relationship (edge, link) between two KG entities. \\
    \underline{type, class}, label, concept & Identifier that represents the same kind or group of entities or relations. \\
    \underline{property (PGM), attribute (RDF)} & an entity feature identifier pointing to a value \\
    \underline{property value}, literal, attribute value & any value that is not referable to as an entity. \\
    \hline
    \end{tabular}
\end{adjustbox}
Furthermore, we refer to the smallest unit of information as \textit{statement} or \textit{fact}. For RDF this would describe a triple, for PGM this can be assigning a property(-value), adding a type label to an entity or adding a relation between two nodes.

\subsection{Requirements of KG construction}
\label{sec:requirements}

\iffalse
- temporal aspects: timestamps and validity
\fi

The development and maintenance of KGs encompass several steps to integrate relevant input data from different sources. 
While the specific steps depend on the input data to be integrated and the intended usage forms of the KG, it is generally desirable that the steps are executed within pipelines with only a minimum of manual interaction and curation. However, a
completely automatic KG construction is not yet in reach since several steps (e.g., identification of relevant sources, development of the KG ontology), as we will see, typically require human input, either by individuals, expert groups or entire communities~\cite{cudre2020leveraging/xi}. 

The KG construction process should result in a high-quality KG based on an expressive \textit{KG data model} as discussed above. The quality of a KG (and data sources) can be measured along several dimensions such as correctness, freshness, comprehensiveness and succinctness~\cite{DBLP:journals/jdiq/MadnickWL009,zaveri2016quality}. 
The correctness aspect is crucial to the validity of information (accuracy) and implies that each entity, concept, relation, and property is \textit{canonicalized} by having a unique 
identifier and being included exactly once (consistency)~\cite{Weikum2021MachineKC}. The freshness (timeliness) aspect requires continuously updating the instances and ontological information in a KG to incorporate all relevant data source changes. The comprehensiveness requirement asks for good coverage of all relevant data (completeness) and that complementing data from different sources is combined~\cite{zaveri2016quality}. Finally, the succinctness criterion asks for a high focus of the data (e.g., on a single domain)~\cite{hogan2021knowledge} and the exclusion of unnecessary information, which also improves resource consumption and scalability of the system (availability). A knowledge graph that meets high standards in these areas can be considered a confident and reliable resource (trustworthiness)~\cite{Wang2021KnowledgeGQ}.

In the following, we discuss requirements for KG construction and maintenance in more detail as they should guide the realization of suitable implementation approaches. 
We group these requirements into four aspects related to 1) input consumption, 2) incremental data processing capabilities, 3) tooling/pipelining, and 4) quality assurance, whereas some essential prerequisites can affect multiple parts of the workflow (e.g., supportive metadata). 
Please note, that we outline the desired functionality for defining arbitrary KG pipelines and that only a subset of it is typically needed for a specific KG project.  
\begin{itemize}
\item \textbf{Input Data requirements}. It should be possible to integrate a large number of data sources as well as a high amount of data (data scalability). There should also be support for heterogeneous and potentially low-quality input data of different kinds such as structured, semi-structured and multimodal unstructured data (textual documents, web data, images, videos, etc.). As a result, KG construction requires scalable methods for the acquisition, transformation, and integration of these diverse kinds of input data. 
The processing of semi-structured and unstructured data introduces the need for knowledge extraction methods to determine structured entities and their relations as well as their transformation into the KG graph data model. Data integration and canonicalization involve methods to determine corresponding or matching entities (entity linking, entity resolution) and their combination into a single representation (entity fusion) as well as matching and merging ontology concepts and properties. For incremental KG construction, the input is not limited to the new data to be added but also includes the current version of the KG and reusable data artifacts such as previously determined mappings specifying how to transform input data into the format of the KG graph model. 

\item \textbf{Support for incremental KG updates}. It should be possible to process the input data both in a batch-like mode where all (new) input data is processed at the same time or in a streaming manner where new data items can continuously be ingested.
The initial version of the KG is typically created in a batch-like manner, e.g., by transforming a single data source or by integrating several data sources into an initial KG. After the initial KG version has been established, it is necessary that the KG can be updated to incorporate additional sources and information.
A simple approach would perform these updates by a complete recomputation of the KG with the changed input data similar than for the creation of the initial KG. However, such an approach would result in an enormous amount of redundant computation to repeatedly extract and transform the same (unchanged) data and to perform data integration and removal of inconsistencies again, possibly with repeated manual interactions. These problems increase with the number and size of input sources and thus limit or prevent scalability. Hence we require support for incremental KG updates that can either periodically be performed in a batch-like manner or in a more dynamic, streaming-like fashion. The batch approach would not require completely rebuilding the KG, but focus on adding the new information without reprocessing previously integrated data. 
A given KG can also be continuously updated with new data in a streaming manner to always provide the most current information for high data freshness. Batch and stream-oriented updates may also be applied in combination~\cite{ilyas2022saga}. As a result, several pipelines may be needed for the creation of the initial KG, the integration of sources with heterogeneous structures, and different forms of incremental KG maintenance.
While  incremental KG maintenance is important in general, specific KG use cases such as research projects may only need a one-time or batch creation of a KG. Hence, the posed requirement would not apply in such cases.  

\item \textbf{Pipeline and Tool Requirements}.
It should be easy to define and run powerful, efficient, and scalable pipelines for creating and incrementally updating a KG. This requires a set of suitable methods or tools for the different steps (discussed in the next section) that should have good interoperability, and a good degree of automation, but still support high customizability, and adapt to new domain requirements.
While the usage of a uniform KG data model (or serialization) can lower debugging complexity of the workflow, reusing existing toolsets might require transformation/mapping between data formats and the processing steps.
Moreover, a pipeline tool should be provided that can integrate the different tools and manages intermediate results and common metadata, e.g., about provenance. The pipeline tool should further provide administration functionality to design and execute pipelines, to support error handling, performance monitoring and tuning etc. Pipeline processes should scale horizontally as new input data is ingested and the KG size increases over time.
Modular processing workflows with transparent interfaces can increase the reusability of alternative tools (implementations).

\item \textbf{Quality Assurance}.
Quality assurance is a cross-cutting topic playing an important role throughout the whole KG construction process. Quality problems in the KG can be multi-faceted relating to the ontological consistency, the data quality of entities and relations (comprehensiveness), or domain coverage. The coverage aspect may focus on the inclusion of relevant data and the exclusion of unnecessary data. In some scenarios, the timeliness of data can play a critical role in real-time-oriented use cases. 
If not handled, quality problems might aggravate over time due to the continuous integration of additional data. 
Therefore methods are needed to evaluate the quality of each step of the construction pipeline as well as of the resulting KG. 
A specific quality aspect is to validate the KG's data integrity concerning its underlying semantic structure (ontology).
Another relevant criterion could be to optimize data freshness to guarantee up-to-date results in upstream applications.
Debugging capabilities based on sufficient metadata are helpful to locate the exact points in the construction pipeline where quality problems arise. Methods are then required for fixing or mitigating the detected quality issues by refining and repairing the KG.
\end{itemize}

\section{Construction Tasks}
\label{sec:tasks}

\begin{figure}
\includegraphics[width=\textwidth]{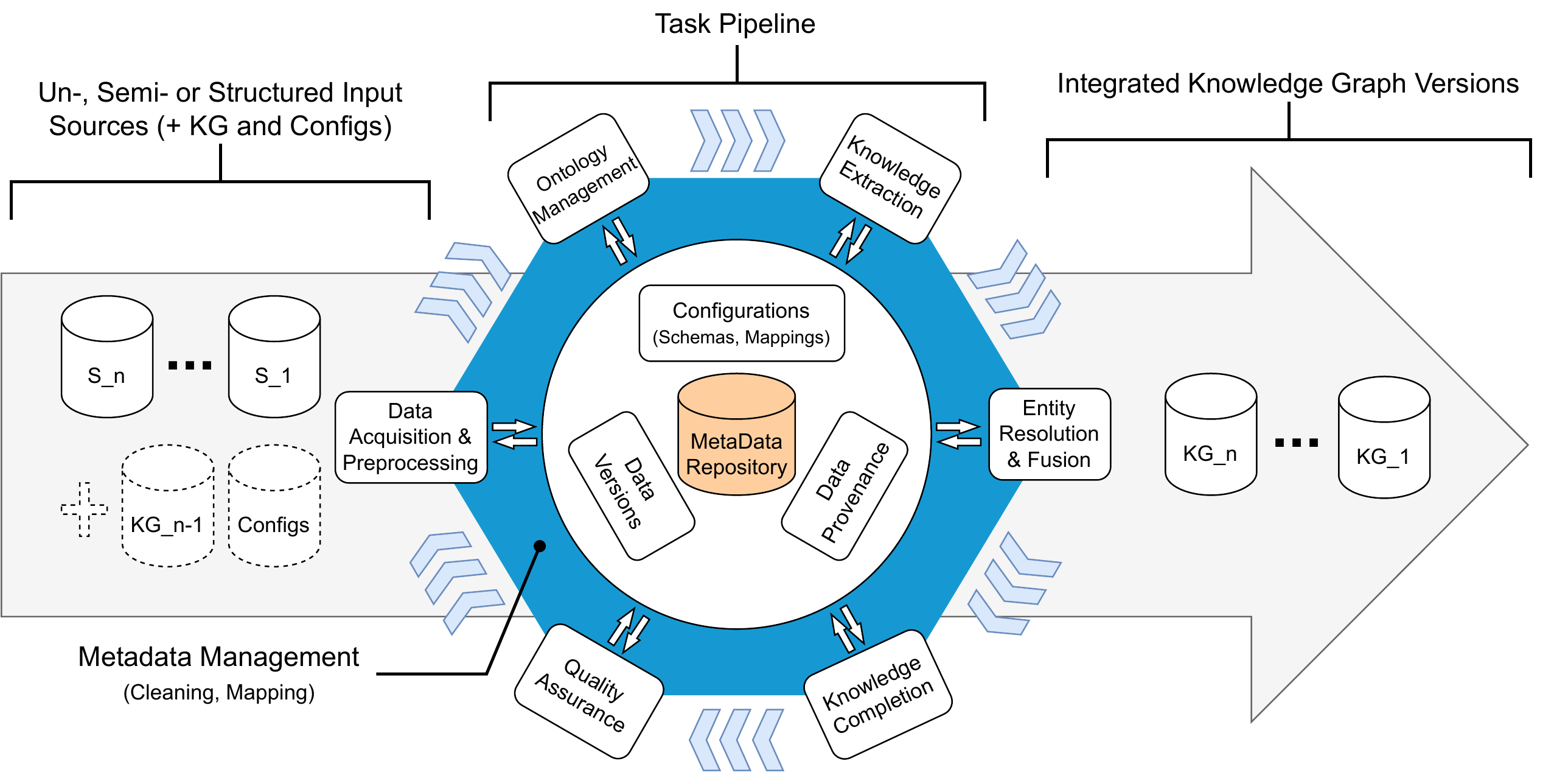}
\caption{Incremental Knowledge Graph Construction Pipeline}
\label{fig:pipeline}
\end{figure}

We give an overview of the main tasks for Knowledge Graph Construction with a focus on (semi-)automatic and incremental solutions. In particular, we cover the following tasks that often involve several subtasks: 
\begin{itemize}
\item Data Acquisition \& Preprocessing:  Selection of relevant sources, acquisition and transformation of relevant source data, initial data cleaning. 
\item Metadata Management: Acquisition and management of different kinds of metadata, e.g., about the provenance of entities, structural metadata, temporal information, quality reports or process logs.
\item Ontology Management: Creation and incremental evolution of a KG ontology.
\item Knowledge Extraction (KE): Derivation of structured information and knowledge from unstructured or semi-structured data.
using techniques for named entity recognition, entity linking and relation extraction. If necessary this also entails canonicalization of entity and relation identifiers.
\item Entity Resolution (ER) and Fusion: Identification of matching entities and their fusion within the KG.
\item Quality Assurance (QA): Possible quality aspects, their identification, and repair strategies of data quality problems in the KG.
\item Knowledge Completion: Extending a given KG, e.g., by learning missing type information, predicting new relations, and enhancing domain-specific data (polishing).  
\end {itemize}

Figure~\ref{fig:pipeline} illustrates a generic pipeline to incrementally incorporate updates from several sources into a KG that may result in  a sequence of distinct KG versions.  It is important to note, that a construction pipeline does not necessarily  follow a fixed execution order for the individual tasks  and  that not all steps may be required  depending on the KG use case. 
This is also because the required tasks depend on the type of source input. Knowledge extraction is commonly applied on unstructured data inputs like text and may not be needed
for structured data, e.g. from databases or other knowledge graphs. Furthermore, the entity linking part of knowledge extraction can make an additional entity resolution step unnecessary. As a result, there may be  different KG construction pipelines for different use cases and data sources. 
The steps of Quality Assurance and KG completion to improve the current version of the KG are not needed for every KG update but may be executed asynchronously, e.g., within separate pipelines (although QA actions such as data cleaning also apply to individual tasks).
Furthermore, data and metadata management play a special role compared to the other tasks, since they are necessary throughout the entire pipeline therefore representing a cross-cutting task, as indicated by the central position of metadata management in Figure~\ref{fig:pipeline}.

\subsection{Data Acquisition \& Preprocessing}
\label{sec:pre-processing}
\subsubsection{Source Selection \& Filtering}

In order to integrate data into a KG, relevant sources must first be identified. Furthermore, relevant subsets of a data source have to be determined as it is generally unnecessary to integrate all information of a source for a given KG project. For example, for a pandemic-specific KG only health-related parts of existing KGs such as DBpedia may be needed.
If the system is not supposed to integrate data sources in their entirety, it can determine a relevant subset by the quality or trustworthiness of a source~\cite{dong2015knowledge} as well as the importance of single entities~\cite{ilyas2022saga}.
Formulating the quantification~\cite{amsterdamer2021automated} of all these criteria alongside computing the cost of integrating a source leads to saving a considerable amount of unnecessary effort while producing a high-quality KG.

Selecting relevant data sources and their subsets are typically manual steps but can be supported by data catalogs providing describing metadata about sources and their contents.
Common approaches to determine such metadata are to employ techniques for data profiling, topic modeling, keyword tagging, and categorization~\cite{fetahu2014scalable,blei2007correlated}.
In~\cite{Weikum2021MachineKC} it is recommended to start KG construction with large curated ("premium") data sources such as Wikipedia and other KGs such as DBpedia. Then further data sources should be identified and integrated to cover additional entities and their relations, especially rather special entities in the "long tail" (e.g., less prominent persons).
Given that sources can differ enormously in size and quality, the order in which sources (and their updates) are integrated can have a strong influence on the final quality~\cite{nentwig2018incremental,saeedi2020incremental,hertling2021order}. Especially for creating the initial KG, these choices are often crucial.
To limit these effects it is advisable to first integrate the sources of the highest quality~\cite{nentwig2018incremental} such as the mentioned premium sources. 
Nevertheless, ideally integration order should not matter in a high-quality pipeline.

\subsubsection{Data Acquisition}
\label{subsubsec:DataAqc}
The data sources of a KG may come in many different data formats such as CSV, XML, JSON, or RDF to meet the requirement of different originating environments and applications.
Furthermore, there are different technologies to exchange or acquire data artifacts by providing downloadable files, deploying databases, or individual application program interfaces (APIs). % supporting access interfaces (APIs). 
Hence, KG construction has to deal with these heterogeneous data formats and access technologies to acquire the data to be integrated.
A common access approach is the use of an adapter component for each source dataset. 
Such an adapter approach is typical for data integration and there are also supporting tools for use in KG management~\cite{DBLP:journals/computer/GieseSVWHJLRXOR15,DBLP:journals/pvldb/CiviliCGLLLMPRRSS13,DBLP:conf/semweb/MamiGSJA019a}.

In addition, KG construction has to deal with continuously changing sources, which necessitates the recognition of such changes and possibly maintaining snapshots of already acquired versions of source data. 
Possible solutions for change detection include manual user notifications over email, accessing a change API using publish-subscribe protocols~\cite{banavar1999efficient}, or computing diffs by repeatedly crawling external data and comparing it with a previously obtained snapshot.   
    
For RDF stores several strategies for maintaining versions of extracted data have been proposed. With full materialization, complete versions (snapshots) of source data are maintained~\cite{Voelkel:2006:SemVersion}. With the delta-based strategy only one full version of the dataset needs to be stored and for each new version only the set of changes or deltas has to be kept~\cite{journals/ijseke/ImLK12,DBLP:conf/i-semantics/GraubeHU14,DBLP:conf/www/SandeCVCMW13,DBLP:journals/ijseke/ImLK12}. The annotated triples strategy is based on the idea of augmenting each triple with its temporal validity~\cite{10.14778/1920841.1920877}. Hybrid strategies have also been considered~\cite{DBLP:conf/er/StefanidisCF14,DBLP:journals/semweb/TaelmanCMV19}. 
Another approach to synchronize changes in a data source 
are Linked Data Event Streams (LDES)~\cite{DBLP:conf/icwe/LanckerCDVMDMBC21}. 
Van Assche et. al.~\cite{DBLP:conf/esws/AsscheORC22} use LDES to continuously update a KG with changes from the underlying data sources.

Other KGs are important sources for data acquisition. However, only a limited number of KGs provide a queryable interface and such interfaces can be expensive to host at high availability~\cite{DBLP:journals/corr/abs-2002-09172,DBLP:conf/semweb/PolleresKFTM18,DBLP:journals/ws/VerborghSHHVMHC16}. To address this problem, recent proposals suggest decentralization, either of the data itself or of the query processing tasks. Decentralization (distribition) of the data across multiple sources~\cite{DBLP:conf/esws/AebeloeMH19,DBLP:conf/semweb/AebeloeMH19,DBLP:conf/www/CaiF04} can increase its availability but tends to provide less efficient query processing  compared to centralized servers or approaches that provide full data dumps to powerful clients for local processing. 
Alternatively, recent studies~\cite{DBLP:journals/corr/abs-2002-09172,DBLP:conf/www/AzzamFABP20,DBLP:conf/otm/HartigA16,DBLP:conf/f-ic/MinierSM19,DBLP:conf/semweb/MontoyaAH18a} have focused on decentralizing the query processing tasks by dividing the processing workload between servers and clients. WiseKG is a system to dynamically distribute the load between servers and clients based on a  cost model~\cite{DBLP:conf/www/AzzamAMKPH21} .

\subsubsection{Transformation \& Mapping}
\label{subsubsec:transformationmapping}
A KG construction pipeline has to transform the input data into the final KG data format, such as RDF or a property graph format. Furthermore, the different pipeline steps may consume and produce different formats, so additional data format transformations or conversions may become necessary. For example, knowledge extraction methods typically process document data such as HTML or Unicode-encoded text, while an entity resolution task may require input data in CSV or JSON format. 

Data transformations have especially been addressed for (semi-)structured data and many tools exist for this purpose~\cite{Junior2016FunULAM}. 
Depending on the required input format, the transformation can be done automatically using generic approaches or requires the manual specification of mappings.
Mapping languages allow the specification of complex and reusable mappings, for example, to transform relational databases (RDB) into an equivalent RDF representation, e.g. using the R2RML language~\cite{dimou2020r2rml}. 
RML~\cite{dimou2014rdf} (RDF Mapping Language) extends R2RML and allows defining mappings not only from RDB but also from other semi-structured data formats such as XML, TSV/CSV, and JSON. 
Systems implementing such mapping languages include SDM-RDFizer~\cite{Iglesias2020SDMRDFizerAR} for RML, and Karma~\cite{KnoblockKarma2012} for an  R2RML alternative called  K2RML.
Relatively little work has so far investigated the transformation of structured data into property graphs~\cite{jain2013graphbuilder,kricke2019graph} although the conversion between RDF and property graphs has received some attention~\cite{angles2020mapping,ReconciliationRDFPG,Abuoda2022TransformingRT}.
In case of an existing RDF-based KG as input, a simple solution for RDF to RDF mappings is to use SPARQL-CONSTRUCT\footnote{\url{https://www.w3.org/TR/rdf-sparql-query/\#construct}} queries, which return a single RDF graph by substituting variables of a given graph pattern with the results of the SPARQL query. As an extension of SPARQL, SPARQL-Generate supports the transformation of streaming and binary data sources~\cite{Lefranois2017ASE}.
The graph query language GQL will support a similar feature for the PGM~\cite{DBLP:conf/sigmod/DeutschFGHLLLMM22}.
An extensive survey on state-of-the-art RDF mapping languages for schema transformation, data transformation, and systems was done by Van Assche et al~\cite{DeclarativeAsscheReview23}. 
A major issue the authors point out is the lack of tools supporting the (semi-)automatic definition of mappings. In their survey, only 3 out of 30 analyzed systems support the semi-automatic definition of mappings (including human-in-the-loop methods)~\cite{Medeiros2015MIRRORAR,Sicilia2016AutoMap4OBDAAG,JimnezRuiz2015BootOXPM}.

\subsubsection{Data Cleaning}
\label{sec:datacleaning}
Data cleaning deals with detecting and removing errors and inconsistencies from data in order to improve the quality of data. Whenever possible, data quality problems within the input sources should be handled already during the import process to avoid that wrong or low-quality data is added to the KG. Data cleaning has received a large amount of interest, especially for structured data, in both industry and research and there are numerous surveys and books about the topic, e.g. \cite{rahm2000data,DBLP:journals/pvldb/AbedjanCDFIOPST16,ilyas2019data,10.1007/978-3-030-71903-6_9}. 

Various types of data errors and quality problems need to be handled ranging from structural to semantic data  problems.  
Dealing with structural problems asks for consolidating  different data structures and formats and ensuring  consistent naming conventions for entities and attributes. For instance, if "USA," "United States," and "U.S." are all used to represent the same country, they should be standardized to a single form. Semantic data cleaning focuses on addressing issues related to the meaning and relationships within the data. One example is handling conflicting or contradictory information present in the dataset. For instance, if one source indicates that a person was born in 1980, while another source suggests 1985, this inconsistency needs to be resolved. Another example is handling entities either in one source or different sources that represent the same real-world object, e.g., a certain customer or product. 

Data cleaning typically involves several subtasks to address these problems. These include data profiling to identify quality issues~\cite{abedjan2018data}, data repair to correct identified problems, data transformation to standardize data representations, and data deduplication to eliminate duplicate entities. Outlier detection is an important aspect of data profiling, aiming to identify data errors based on the assumption of specific "normal" data values. For instance, it is highly unlikely that an individual born in the mid-19th century would still be alive in the year 2020.

Rule-based methods are classic techniques used for data cleaning. These methods handle errors that violate integrity constraint rules, such as functional dependencies (FDs)~\cite{DBLP:journals/pvldb/BeskalesIG10,DBLP:conf/icde/BeskalesIGG13,DBLP:conf/sigmod/KhayyatIJMOPQ0Y15,DBLP:conf/icdt/KolahiL09,DBLP:conf/sigmod/WangH19}, conditional functional dependencies (CFDs)~\cite{DBLP:conf/icde/BohannonFGJK07,DBLP:journals/tods/FanGJK08,DBLP:journals/pvldb/GeertsMPS13,DBLP:conf/sigmod/KhayyatIJMOPQ0Y15}, and denial constraints (DCs)~\cite{DBLP:conf/icde/ChuIP13,DBLP:journals/pvldb/GeertsMPS13,DBLP:conf/sigmod/HeidariMIR19,DBLP:conf/sigmod/KhayyatIJMOPQ0Y15,DBLP:conf/icde/LopatenkoB07,HoloClean}. While rule-based methods can handle data that violates predefined rules, their effectiveness is limited by the challenge of obtaining sufficient and correct rules.
Statistical cleaning methods repair errors based on probabilistic distributions within the data~\cite{DBLP:conf/sigmod/HeidariMIR19,DBLP:journals/pvldb/KrishnanWWFG16,DBLP:conf/sigmod/MahdaviAFMOS019,DBLP:conf/icde/ZhengMC19,HoloClean,DBLP:conf/sigmod/WangH19}.

User interaction cleaning methods involve human knowledge to enhance the quality of cleaning results while minimizing the effort required~\cite{DBLP:conf/icde/AssadiMN17,DBLP:journals/pvldb/ChuOMIP0Y15,DBLP:conf/sigmod/HeVSLMPT16,DBLP:journals/pvldb/KrishnanWWFG16,DBLP:conf/sigmod/MahdaviAFMOS019,DBLP:conf/sigmod/Thirumuruganathan17,DBLP:conf/icde/TongCZLC14,DBLP:journals/pvldb/YakoutENOI11}. The use of machine learning for data cleaning has gained prominence in recent years, as it simplifies the configuration of various subtasks. For example, HoloClean~\cite{HoloClean} employs observed data to build a probabilistic model for predicting unknown data values. Other applications of machine learning for data cleaning are covered in~\cite{ilyas2019data,neutatz2021cleaning}.

If there is already a KG version to be extended, the KG information can be leveraged to identify and handle data errors. For instance, KATARA~\cite{DBLP:journals/pvldb/ChuOMIP0Y15} employs crowdsourcing to verify whether values that don't match the KG are correct or not. Hao et al.~\cite{DBLP:journals/vldb/HaoTLLF18} introduce detective rules (DRs) that can make actionable decisions on relational data, by building connections between a relation and a KG. KGClean~\cite{DBLP:journals/corr/abs-2004-14478} is an initial attempt at a KG-driven cleaning framework utilizing knowledge graph embeddings.

Techniques for ensuring KG quality are discussed in Section~\ref{sec:evaluation}. Approaches for identifying duplicates across data sources (entity matching) are outlined in Section~\ref{sec:entity-resolution}. It is advantageous to remove duplicates within a source early on to simplify the deduplication process across sources.

\subsection{Metadata Management}
\label{sec:metadata-management}
Metadata describes data artifacts and is important 
for the findability, accessibility, interoperability and \mbox{(re-)usability} of these artifacts~\cite{wilkinson2016fair,kricke2017preserving,Ma2021KnowledgeGC}.
There are many kinds of metadata in KGs such as descriptive metadata (content information for discovery), structural metadata (e.g. schemas and ontologies), and administrative metadata concerning technical and process aspects (e.g., provenance information, mapping specifications)~\cite{greenberg2005understanding,neto2017idol,duval2002metadata}.  
It is thus important that KG construction supports the comprehensive representation, management and usability of the different kinds of metadata. From the perspective of KG construction pipelines, this includes metadata for each data source (schema, access specifications), each processing step in the pipeline (inputs including configuration, outputs including log files and reports), about intermediate results and of course the KG and its versions. Moreover, for each fact (entity, relation, property) in the KG there can be metadata such as about provenance, i.e., information about the origin of data artifacts. Such fact-level provenance is sometimes called \textit{deep} or \textit{statement-level provenance}. Examples of deep provenance include information about the creation date, confidence score (of the extraction method) or the original text paragraph the fact was derived from.
Such provenance can help to make fact-level changes in the KG without re-computing each step or to identify how and from where wrong values were introduced into the KG~\cite{kricke2017preserving}.

Metadata can be created either manually by human users (e.g., to specify a license for data usage or a configuration of a pipeline step) or by a computer program based on a heuristic or an algorithm~\cite{neto2017idol}.
In the latter case the results may be exact or only approximate. For example, data profiling computes accurate statistical information (e.g., about the distribution of values) while the use of machine learning (e.g., for type recognition) usually does not provide perfect accuracy. %Analytical metadata contains exact results like statistical metadata.

To use metadata effectively for KG construction, it is beneficial to maintain a metadata repository (MDR)  to store and organize the different kinds of metadata in a uniform and consistent way~\cite{wilkinson2016fair}.
The MDR can either be separate from the sources and the KG with references to data artifacts or there can be combined solutions for both the data and their metadata.  
While there may be several metadata repositories for the different sources and processing steps, a central solution can simplify access to all KG-relevant metadata.
In contrast, using multiple metadata solutions might allow more flexibility in selecting specialized solutions that suit specific needs or types of metadata. This approach can also introduce complexity or inconsistencies and hinder the process of discovery and exploration due to information being scattered across various repositories.
Specific implementations of MDRs are CKAN\footnote{\url{https://ckan.org/}}, Samply~\cite{kadioglu2018samply}, or the DBpedia Databus~\cite{frey2022managing}, all using specific vocabularies, standard query languages, and databases to implement their relevant features. 
Concerning metadata exchange, the Open Archives Protocol for Metadata Harvesting\footnote{\url{ https://www.openarchives.org/OAI/openarchivesprotocol.html}} framework also allows acquisition of structured metadata. %acquisition using a client-server approach. 

Fact-level metadata (or annotations) in the KG can be stored either together with the data items (embedded metadata) or in parallel to the data and referenced using unique IDs (associated metadata)~\cite{duval2002metadata}.
For example, fact-level metadata can support the selection of values and sub-graphs~\cite{frey2019dbpedia}, or the compliance to used licenses in target applications.
Such annotations are also useful for other kinds of metadata.
Temporal KGs can be realized by temporal annotations to record the validity time interval (period during which a fact was valid) and transaction time (time when a fact was added or changed)~\cite{rost2021bitemporal,rost2021distributed}. The possible implementations for fact-level annotations depend on the used graph data model (see Section~\ref{sec:models}). 
From another perspective, provenance metadata can also capture the steps of the applied schema and data transformations in the pipeline~\cite{Meester2017DetailedPC,Meester2020ImplementationindependentFR}.

\subsection{Ontology Management}
\label{sec:ontology-development}

Ontology development is the incremental process of creating or extending an ontological knowledge base~\cite{noy2001ontology}.
KG construction requires to develop an ontology for the initial KG and to incrementally update the ontology to incorporate new kinds of information. 
As of today, 
ontology development and curation is still done broadly manual or crowdsourced although some semi-automatic approaches are also proposed. 
Semi-automatic ontology development tasks share a great overlap with methods from knowledge extraction, entity resolution, quality assurance and knowledge completion. 
Creating the initial ontology can be derived from a single source that ideally provides already some useful ontology to build on. 
Public web wikis, catalogs, APIs, or crowdsourced databases are valuable starting sources as they may already contain a large amount of (semi-)structured data on general or domain-specific topics. However, cleaning and enrichment processes are required to ensure sufficient domain coverage and quality to build an initial knowledge graph structure from this existing data.
For example, if Wikipedia is used as a primary source its category system can be good start to derive the most relevant classes for the KG by some NLP-based "category cleaning"~\cite{Weikum2021MachineKC}. 
Semi-automatic approaches mostly focus on learning an ontology from single sources, i.e. transforming a source into an ontology or KG. These individual ontologies or KGs can then be integrated into a previous version of the overall KG. A key prerequisite for this kind of ontology integration is the step of ontology and schema matching to determine respective ontology and schema elements (classes, properties). 
After a discussion of semi-automatic approaches for ontology learning we therefore discuss ontology/schema matching and close with approaches for ontology integration.

\subsubsection{Ontology Learning}
There are two main subfields of ontology learning; the first focuses on learning from text sources (unstructured data); the second from relational databases (structured data).
Although the authors in~\cite{al2020automatic,browarnik2015ontology,wong2012ontology} discuss that automatic ontology construction is not likely to be possible, a significant amount of research has been done to support the semi-automatic construction for single sources.

Al-Aswadi et. al.~\cite{al2020automatic} give a state-of-the-art overview of ontology learning from \textbf{unstructured text} where the goal is to identify the main concepts and their relations for the entities in a document collection. The approaches~\cite{wong2012ontology,al2020automatic,browarnik2015ontology} can be grouped into linguistic approaches (using NLP techniques such as part-of-speech tagging, sentence parsing, syntactic structure analysis, and dependency analysis methods) and machine learning approaches. The latter 
include statistic-based methods (e.g., utilizing co-occurrence analysis, association rules, and clustering) and 
logic-based approaches using either inductive logic programming or logical inference.
Al-Aswadi et. al.~\cite{al2020automatic} argue that there is a need to move from shallow to deep learning approaches for deeper sentence analysis and improved learning of concepts and relations. 

Ma et al.~\cite{ma2022ontology} give a survey of methods for learning ontologies from \textbf{relational databases} with a focus of methods for reverse engineering or the use of mappings to transform a relational database (schema) into an ontology or knowledge graph. 
Reverse-engineering allows one to derive an Entity-Relationship diagram or conceptual model from the relational schema. Here additional considerations are needed to deal with trigger and constraint definitions to avoid a semantic loss in the transformation.
For the mapping techniques to transform RDBs to KGs, the authors differentiate between template-based, pattern-based, assertion-based, graph-based mapping, and rule-based mapping approaches. % RDF or RDBs to OWL.
The resulting mappings should be executable on instance data in order to generate a graph structure from a relational database~\cite{de2014r2g,petermann2014biiig}. 

\subsubsection{Ontology/schema matching}
Consolidating and integrating information from multiple heterogeneous sources requires harmonizing the ontologies and/or schemas of the sources. A main step for such a data integration is ontology and schema matching which is the task of identifying corresponding ontology and schema elements, i.e. matching ontology concepts and matching properties of concepts and entities. For example, to integrate a new source into a KG it is necessary to perform a matching of the source ontology/schema with the KG ontology to identify which source elements are already existing in the KG ontology and which ones should be added. Property matching is also important for entity resolution and entity fusion in order to determine matching entities based on the similarity of equivalent properties and to combine equivalent properties to avoid redundant information. In some cases known entity matches can be used to aid in the ontology matching step~\cite{DI2KGFAMER}. Some tools also perform entity resolution and ontology matching in combination~\cite{SuchanekPARIS}.

There is a huge amount of previous research on schema and ontology matching, although mostly outside the context of knowledge graphs, and there are numerous surveys and books about the topic~\cite{rahm2001survey,euzenat2007ontology,zohra2011schema,Rahm2011,Otero-Cerdeira2015}. The match approaches typically rely on determining the similarity of elements using different strategies, such as the similarity of concept/property names or the similarity of instance values. 
Structural information can also be beneficial in the matching process, e.g. by looking at the concepts in the graph neighborhood. Matching systems commonly rely on a combination of different match strategies in order to achieve high-level match-quality~\cite{do2002coma}.

While string similarity can be a strong signal for a match decision, often semantically similar words are used, which are dissimilar on character level. Dictionaries or pre-trained word embeddings are therefore helpful to capture this semantic similarity.
Zhang et. al.~\cite{Zhan2014OntMatchWordEmb} investigate how word embeddings (using Word2Vec~\cite{word2vec}) can be used for the task of ontology matching. 
They found a hybrid approach to perform the best, which takes the maximum of either edit-distance-based or word embedding similarity for each entity pair. 
Another approach, which relies on word embeddings but also on meta-information about property's names and their values as input for a dense neural network is LEAPME~\cite{ayala2020leapme}.
This system trains a classifier based on already labeled property pairs.
This trained model can then decide whether unlabeled property pairs and their similarity scores constitute a match.
Graph embeddings have also seen some attention in ontology matching as they can capture structural information of an ontology. For example Portisch et. al.~\cite{OntMatchAbsOrient} use a variation of RDF2Vec~\cite{RDF2VecLight}, which is a walk-based embedding technique similar to Word2Vec, to encode both ontologies and then use a rotation matrix to align the embeddings.  

There are some tailored ontology matching approaches for KGs such as for mapping categories derived from Wikipedia to the Wordnet taxonomy with the goal to achieve an enriched KG ontology~\cite{Weikum2021MachineKC}.
Other specific approaches have been developed to address the integration of RDBs with ontologies, that go beyond the mapping languages described in Section~\ref{subsubsec:transformationmapping}. KARMA~\cite{KnoblockKarma2012} provides a semi-automatic approach to link a structured source such as a RDB with an existing ontology. The process consists of assigning semantic types to each column, constructing a graph of all possible mappings between the source and the ontology, refining the model based on user input and finally generating a formal specification of the source model.

\subsubsection{Ontology Integration}

\begin{figure}[!h]
    \centering
    \includegraphics[width=\textwidth]{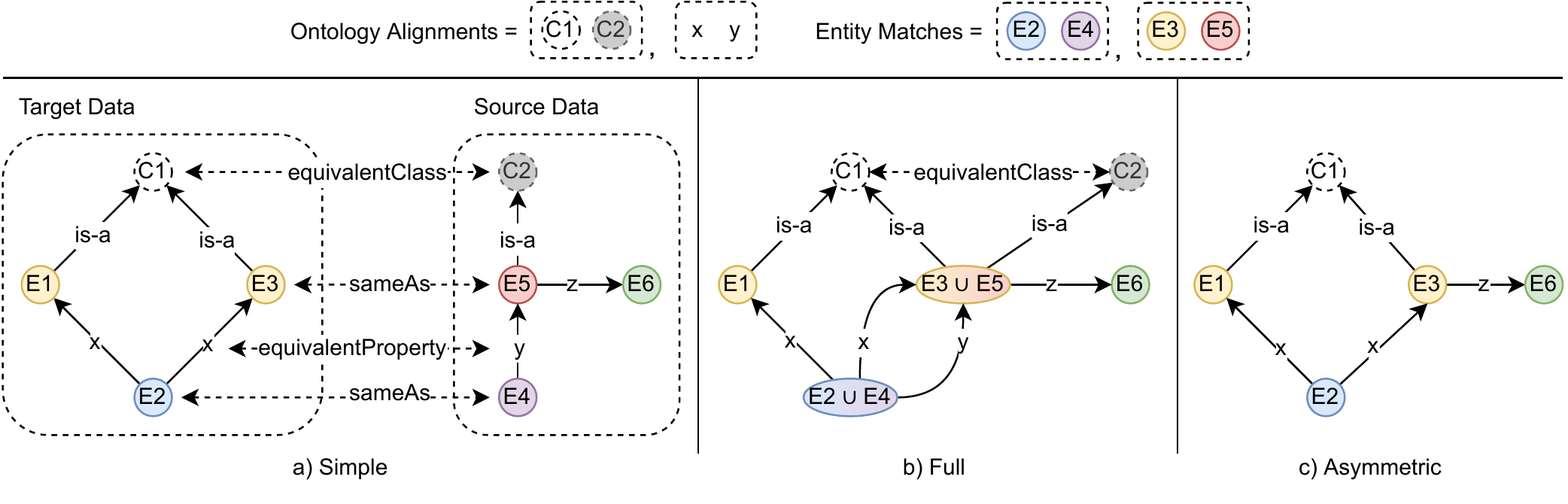}
    \caption{Ontology and Entity Merging Strategies.}
    \label{fig:merge}
\end{figure}

Merging new ontology or schema data into the existing KG ontology is a subtask of ontology or schema integration.
This topic has achieved some attention where recent approaches utilize the mapping result of a match (alignments) to combine multiple ontologies/schemas~\cite{pottinger2003merging,raunich2014target,osman2021ontology}. If the match mapping is automatically determined, it must first be manually validated and possibly corrected to provide a valid basis for the merge step. 
Osman et al.~\cite{osman2021ontology} give a comprehensive and recent summary of ontology integration techniques, which can handle the merging of ontology and entity data using respective alignments.
The authors distinguish the following merging strategies:
\begin{itemize}
\item \textbf{Simple Merge.} Imports all input ontologies into a new ontology and adds bridging constructs between equivalent entities, like defining OWL \textit{equivalentClass} or \textit{equivalentProperty} relations.
\item \textbf{Full Merge.} Imports all source ontologies into a new ontology and merges each cluster of equivalent entities into a new unique entity with a union of all their relations and leaving equivalent classes untouched.
\item \textbf{Asymmetric Merge.} These approaches import source ontologies into a preferred target ontology, preserving all its concepts, relations, and rules by merging matching entities into existing target entities or else by creating new ones.
\end{itemize}
Figure~\ref{fig:merge} visualizes each of the three strategies, where a) also shows the source and target data that are merged. From the three merging strategies, the authors favor the last and mention it as a good solution for incremental ontology integration. The reason for this preference is that the asymmetric merge strategy preserves the target ontology during the integration, and only adds new elements from a source ontology if necessary. This can also be seen, in Figure~\ref{fig:merge} c), where the target data is left unchanged, and only a new entity \texttt{E6} is added, which is connected to \texttt{E3} from the target data. An example approach for asymmetric ontology merging focusing on is-a  relations  is proposed in~\cite{raunich2014target}.

\subsection{Knowledge Extraction}
\label{sec:knowledge-extraction}
\begin{figure}[!b]
    \centering
    \includegraphics[width=\textwidth]{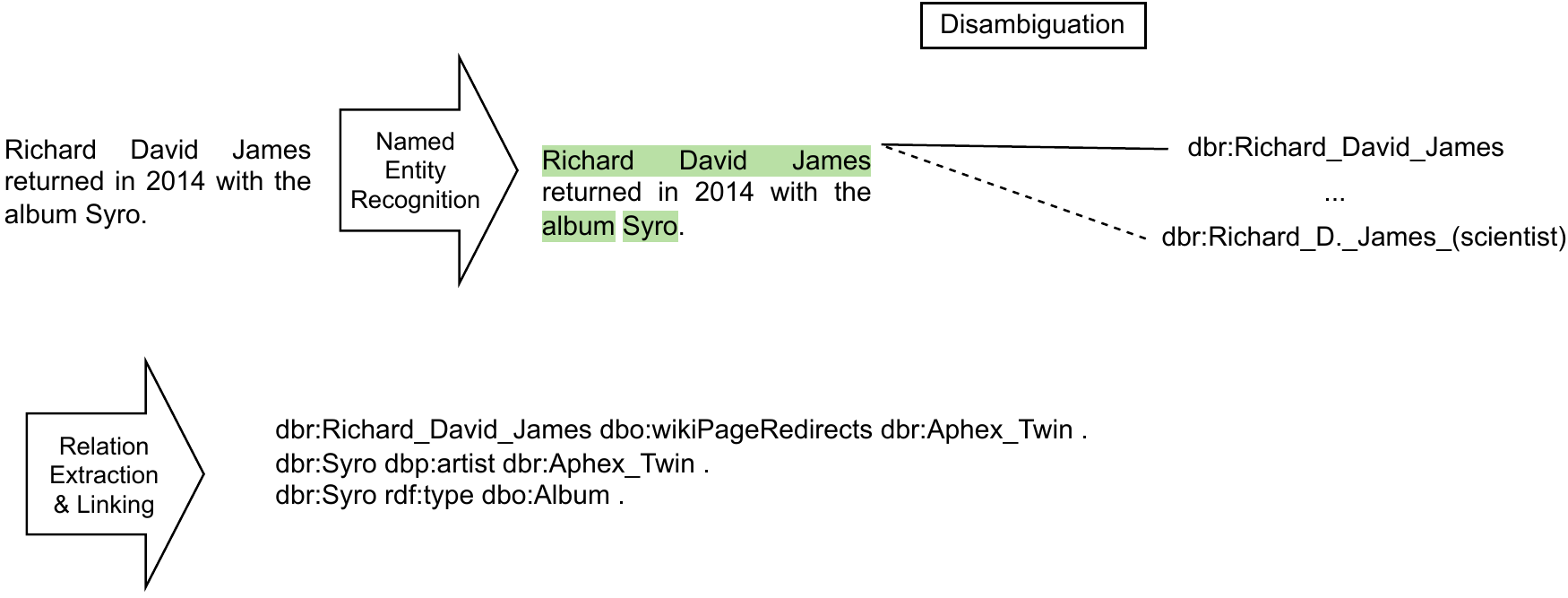}
    \caption{Knowledge Extraction steps for an example sentence linking entities and relations to the DBpedia KG.}
    \label{fig:ke}
\end{figure}

Knowledge extraction is a process to obtain structured, more computer-readable data from unstructured data such as texts or semi-structured data, like web pages and other markup formats.
The extraction methods of semi-structured data often use a combination of data cleaning (Section~\ref{sec:datacleaning}) and rule-based mappings (Section~\ref{subsubsec:transformationmapping}) to transform the input data into the final KG, targeting already defined classes and relations of the existing ontology.
Most of the work focuses on knowledge extraction from text, sometimes additionally considering images and figures within the text. 
Recently, there has been an increased interest in creating multi-modal knowledge graphs (i.e. KGs with not only text but also other modes of data such as images) necessitating appropriate methods of knowledge extraction. The detailed discussion of such methods lies outside the scope this paper, but we refer the interested reader to the following survey by Zhu et. al.~\cite{Zhu2022MultiModalKG}.
The main steps of text-based knowledge representation are named-entity recognition, entity linking, and relation extraction. These steps are discussed in the following and allow the extraction of entities and relations from text for inclusion into a KG. An example of this process is shown in Figure~\ref{fig:ke}.

\subsubsection{Named Entity Recognition}
Named-entity recognition (NER) refers to demarcating the locations of entity mentions in an input text. In the most widely used scenarios mentions of only a handful of types (e.g.~persons, places, locations, etc.) are determined. However KGs usually contain hundreds or thousands of types. Furthermore, off-the-shelf NER tools do not provide canonicalized identifiers for the extracted mentions. A second step is therefore necessary to link entity mentions either to existing entities in a KG or with new identifiers.

A relatively reliable and simple way to detect entity mentions in a text is the use of a \textit{dictionary} (also referred to as \textit{lexicon} or \textit{gazetter}), which maps labels of desired entities to identifiers in the KG. In addition to it's simplicity such an approach already provides recognized entities in a text with the right link to the KG (i.e. solving the tasks of named-entity recognition and entity linking in one step). However these dictionaries are usually incomplete.
A simple way to increase the coverage of such dictionaries is to utilize disambiguated aliases in high-quality sources~\cite{Weikum2021MachineKC}. Wikipedia redirects or DBpedia's \texttt{dbo:alias} property would be a simple way to enhance an entity dictionary.
For example \url{https://en.wikipedia.org/wiki/Richard_D_James} from the example redirects to \url{https://en.wikipedia.org/wiki/Aphex_Twin}.
To make dictionary lookups efficient different data structures have been proposed. For example prefix tries or inverted indexing have shown to be a scalable solution for large Web search engines and are used in the NER approaches AGDISTIS~\cite{AGDISTIS_ISWC}, TagME~\cite{TagMe} and WAT\cite{WAT}.

Machine learning methods have become increasingly popular to tackle NER. This is especially useful to find "emerging entities", i.e. entities that are unknown to the knowledge base.
The machine learning models for entity recognition generally fall into the task of \textit{sequence labeling}. 
A widely successful method for this task is known as Conditional Random Fields (CRF), which uses an undirected graph connecting input and output variables and models the conditional probability of output given the input. Generally these graphs form a linear chain (e.g in the Stanford CoreNLP package~\cite{StanfordCoreNLP}), which means for a prediction only the immediate neighbors are relevant in a sequence.
While CRFs require extensive feature engineering Deep Neural Networks have become highly popular in recent years for the task of NER, since they do not necessitate this amount of human interaction.
For example \textit{LSTM} networks ("long short term memory"), which are a specific case of \textit{recurrent neural networks} (RNN), have become a prevalent choice for NER tasks~\cite{DBLP:series/synthesis/2017Goldberg}. The memory cells contained in this architecture are able to deal with long-term dependencies, which was previously a major painpoint for RNNs.
Deep learning-based approaches for NER are surveyed in~\cite{DeepLearningNER}.

As multi-modal data become increasingly popular, e.g. on social media platforms, several recent studies focus on multi-modal NER (MNER)~\cite{moon-etal-2018-multimodal,yu-etal-2020-improving-multimodal}, where the goal is to leverage the associated images to better identify the named entities contained in the text. Furthermore, there are first approaches~\cite{pezeshkpour-etal-2018-embedding,li-etal-2020-gaia,DBLP:journals/corr/abs-2203-09138} that address MNER for KGs. They aim to correlate visual content with textual facts. One typical solution parses images and texts to structured representations first and grounds events/entities across modalities. However, intra-modal relation extraction and cross-modal entity linking still are largely unresolved problems.

 \subsubsection{Linking}
If named entities are recognized in a text they need to be linked to the knowledge base or KG. This is called \textit{entity linking} (EL) or \textit{named entity disambiguation} (NED). Given a set of candidates from the knowledge base an EL algorithm needs to decide which entity a mention belongs to. In Figure~\ref{fig:ke} this can be seen, where \texttt{Richard David James} is linked to the DBpedia entity \texttt{dbr:Richard\_David\_James}.

EL algorithms can rely on a variety of features. Based on the mention itself the \textit{confidence} of the used NER tool can be used, how \textit{similar} the mention and the entity are or how much \textit{overlap} exists across mentions~\cite{Martinez_Rodriguez_2020}. 
The context of the extracted mentions can be valuable. Keyword-based similarity can be used by relying on TF-IDF scores, where rare keywords used in the mention's context, which connect to a candidate entity, can give hints for linkage~\cite{Kulkarni_2009,Milne_2008}.
Words that occur frequently in the same context can also help in the disambiguation process. Here pre-trained word embeddings can prove especially useful, since they encode semantic similarity in a latent space. 
Furthermore, already disambiguated mentions can be used to aid in the linking of entities that occur in the same paragraph.

\textit{Holistic} entity linking~\cite{DBLP:conf/sigir/HanSZ11,Martinez_Rodriguez_2020} approaches leverage background information in the decision process, that exceeds merely using the similarity between mention and entity.
Popularly, the graph-structure of Wikipedia links can be used to determine \textit{commonness} and \textit{relatedness}. 
Commonness refers to the probability that an entity mention links to the respective Wikipedia article of the given candidate entity.
Relatedness measures how many articles in Wikipedia link to articles of both candidates. 
Using such background knowledge unambiguous mentions can aid in the correct linkage of ambiguous mentions~\cite{Medelyan08topicindexing}.

Entity linking approaches furthermore need to address specific challenges such as \textit{coreference resolution}, where entities are not consistently referred to by their names, but with indirect references such as pronouns~\cite{Martinez_Rodriguez_2020} and how to deal with \textit{emerging entities}, i.e. entities that are recognized, but not yet existing in the target KG.
For example Hoffart et. al~\cite{HoffartEmergingEntitiesWWW2016} keep a contextual profile of emerging entities and when this profile contains enough information to infer the semantic type of the mention it can be added to the KG with its type.

Generally entity linking and the later discussed entity resolution (Section \ref{sec:entity-resolution}) share similarities in aiming to connect the same entities in and across data sources. entity linking and entity resolution are sometimes jointly discussed under the term \textit{entity canonicalization}~\cite{Weikum2021MachineKC}.
While entity resolution typically deals with at least semi-structured data sources, there have been some efforts to address cases with unstructured sources, where deep learning-based approaches are advantageous~\cite{deepmatcher}.
However, there are some key differences not only in the characteristic modality of the data sources, but also in the signals that lead to a linking decision.
For example, in the entity linking scenario, if one has already linked the mention "Richard James" to the entity \texttt{dbr:Richard\_David\_James}, seeing the mention "James" in close context makes it likely that it also refers to the same entity.
By contrast, in the entity resolution scenario, if one already confidently matched two entities, it is unlikely that a similar unmatched entity from one data source will also be matched with the already matched entity from the other data source.
This is because matching can often focus on $1-1$ correspondence between two data sources under the assumption of deduplicated or clean data sources.
However, investigating how well entity resolution approaches for dirty sources (where multiple entities may match with the same KG entity) can be utilized for entity linking and vice versa could be worthwhile.

\subsubsection{Relation Extraction}
Given the identified entities in a text, relation extraction aims to determine the relationship among those entities. 
In Figure~\ref{fig:ke} we see this for example, when the text snippet \texttt{album Syro} becomes the triple \texttt{dbr:Syro rdf:type dbo:Album}, i.e. the type relation for entity \texttt{dbr:Syro} is determined. 

The first techniques use rule-based approaches to extract relations, e.g. by relying on Hearst Patterns to find hyponym (\textit{is-a}) relations~\cite{HearstPatterns92} or involving regex expressions~\cite{SnowballAgichtein2000,DIPREBrin98}. In order to improve coverage different ways to enhance such patterns were devised. The human involvement in these techniques however is a limiting factor.
To address these shortcomings, statistical relation extraction models were devised.
Feature-based methods rely on lexical, syntactic, and semantic features to use as input for relation classifiers.
Similarly, kernel-based methods~\cite{DBLP:conf/emnlp/ZhouZJZ07} rely on specifically designed kernel functions for SVMs to measure the similarity between relation candidates and text fragments.
Graph-based methods further integrate known relations between entities and text in order to correctly identify relations~\cite{Weikum2021MachineKC}.

While such methods can be incredibly useful to obtain relatively simple relations with high accuracy they are limited in terms of their recall or at least require a high degree of additional human involvement for feature engineering, designing of kernel functions or the discovery of relational patterns~\cite{RelationExtractionCNNPerspectiveNguyenG15, PCNNRelationExtractionZeng2015}. 
Neural relation extraction methods aim to close this gap. The input text is transformed via (pre-trained) word embeddings and position embeddings into a format that is suitable for the neural networks that are trained for relation extraction.
Instead of devising hand-crafted features the focus in this area lies on investigating various neural network architectures such as recurrent neural networks, convolutional neural networks and LSTMs.
The bottleneck for these approaches lies in the availability of training data. A common approach to address this is via distant supervision.
Statements from a given data source (for example Wikipedia) are used to train the given model.
Especially the use of pre-trained language models has pushed the state-of-the-art to new heights~\cite{baldini-soares-etal-2019-matching,DBLP:conf/cikm/WuH19a}.
Han et. al~\cite{han-etal-2020-data} provide a more in-depth overview over these methods and identify the main challenges in the ability to utilize more data, creating more efficient learning schemes, handling more complex contexts (e.g. relational information \textit{across} sentences) and detecting undefined relations in new domains.

A special case of relation extraction aims to extract relations freely without a pre-defined set of relations. This is known as Open Information Extraction (OpenIE). While this can be a good way to increase the variety of information contained in the KG, a secondary step is necessary to \textit{canonicalize} the extracted relations in order to deduplicate and possibly even link them to already contained synonymous relations in the KG~\cite{cesi2018}.

Several tools exist for the entire process of Knowledge Extraction, with some tools focusing on specific aspects. For example DBpedia Spotlight~\cite{DBpediaSpotlight} mainly aims at performing named entity extraction and links those mentions to the DBpedia KG. The dstlr~\cite{dstlr} tool extracts mentions and relations from text, links those to Wikidata and furthermore populates the resulting KG with more facts from Wikidata. OpenNRE~\cite{OpenNRE} provides an extensible framework for neural relation extraction, with trainable models, however this approach would necessitate an independent linking step afterwards.

Analogously to NER there are also efforts to use images as information sources for relation extraction. These can range from rule-based approaches~\cite{elliott-keller-2013-image}, which for example verbalize detected spatial relations of recognized entities in an image, to learning-based techniques, which encode visual features of detected objects as well as textual features into distributed vectors used to predict relations between given objects. For example MEGA~\cite{ZhengMEGAMultiModalRE} aligns information contained in the syntax tree and word embeddings of the textual data and the scene graph obtained from the image. A scene graph connects detected objects in an image via their visual relations. After the alignment process the respective representations are concatenated and sent to a Multilayer Perceptron (which is a fully-connected feed-forward neural network) to predict the relation.

\subsection{Entity Resolution and Fusion}
\label{sec:entity-resolution}
Entity resolution (ER), also called entity matching, deduplication or link discovery, is a key step in data integration and for good data quality. It refers to the task of identifying entities either in one source or different sources that represent the same real-word object, e.g., a certain customer or product. 
An enormous amount of research has dealt with the topic as evidenced by numerous surveys and books~\cite{kopcke2010frameworks,christen2012data,nentwig2017sw,barlaug2021neural,christophides2020bigdataER,papadakis2020blocking}. In addition to several research prototypes there are also many commercial solutions such as IBM's InfoSphere Identity Insight\footnote{\url{https://www.ibm.com/products/infosphere-identity-insight}} or SAP's Master Data Governance Platform\footnote{\url{https://www.sap.com/products/technology-platform/master-data-governance.html}}. 
Most known approaches tackle static or batch-like entity resolution where matches are determined within or between datasets of a fixed size. The more recent of these approaches deal with multi-source big data entity resolution~\cite{clip,jedai3}, rely on Deep Learning~\cite{ebraheem2017DLBlk2,deepmatcher} or KG embeddings~\cite{OpenEA,eagerkgcw2021}, with the neural methods having seen more scrutiny recently after an era of relative hype~\cite{CriticalReevalNeuralER}.

For KG construction, however, we need incremental approaches that build on previous match decisions and determine for new entities if they are already represented in the KG or whether they should be added as new entities. Furthermore, for streaming-like data ingestion into a KG a dynamic (real-time) matching of new entities with the existing KG entities should be supported.  
Entity resolution results are fed to the step of \textit{entity fusion} which fuses together matching entities to combine and thus to enrich the information about an entity in an uniform way.  

In the following we first discuss proposed approaches for incremental ER and then for entity fusion. 
 
\subsubsection{Incremental Entity Resolution}
\label{subsubsec:IncER}
Entity resolution is challenging due to the often limited quality and high heterogeneity of different entity descriptions. It is also computationally expensive because the number of comparisons between entities typically grows quadratically with the total number of entities. The standard approach for entity resolution uses a pipeline of three succeeding phases called blocking, linking/matching and clustering~\cite{saeedi2020incremental, PapadakisFourGenER}.
The main step is to determine the similarity between pairs of entities to determine candidates for matching. This matching step often results in a similarity graph where nodes represent entities and edges link similar pairs of entities.
The preceding blocking phase aims at drastically reducing the number of entity pairs to evaluate, e.g. based on some partitioning so that only entities of the same partition need to be compared with each other (e.g., persons with the same birth year or products of the same manufacturer).
After the match phase there is an optional clustering phase that uses the similarity graph to group together all matches. This phase can typically improve the quality of entity resolution, by relying on a more holistic perspective on entity similarities, when compared to the myopic pairwise matching. The clustering step also assists the succeeding step of entity fusion to fuse the matching entities into one representative entity for the KG.

\begin{figure}
    \centering
    \includegraphics[width=\textwidth]{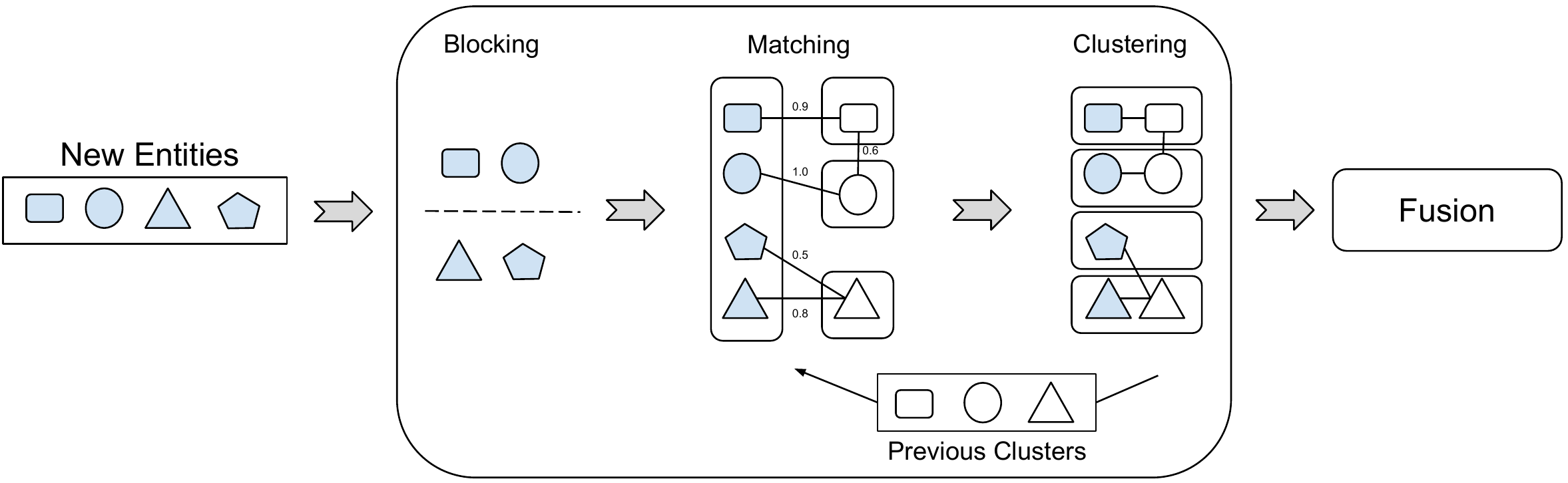}
    \caption{Incremental entity resolution workflow.}
    \label{fig:erWorkflow}
\end{figure}

For incremental ER the task is to match sets of new entities from one or several sources with the current version of the KG which is typically very large and contains entities of different types. It is thus beneficial to know the type of new entities from previous steps in the KG construction pipeline so that only KG entities of the same or related types need to be considered. 
\Cref{fig:erWorkflow} illustrates a high level workflow for incremental ER. 
The input is the current version of the KG with the already integrated entities (previous clusters in 
\Cref{fig:erWorkflow}) as well as the set of new entities to be integrated. This requires the development of incremental versions for blocking, matching and clustering phases that focus on the new entities. To allow better match decisions for incremental ER it is generally advantageous to retain the entities of the previously determined clusters (and their match similarities) and not only the fused cluster representatives. Some incremental clustering schemes can also use this to identify previous match mistakes and to repair existing clusters for new entities~\cite{gruenheid2014vldb,saeedi2020incremental}. Some approaches~\cite{gazzarri2021end} and tools~\cite{saeedi2018csimq} also support to execute incremental ER in parallel on multiple machines to improve execution times and scalability to deal with large KGs.

\textit{Blocking} for incremental or streaming ER requires to identify for the new entities all other entities in the KG that need to be considered for matching. Given the typically high and growing size of the KG it is important to limit the matching to as few candidates as possible and determining the candidates should also be fast.  As mentioned, blocking and entity resolution should be limited to entities of the same (or most similar) entity type and one can apply the same blocking approach for the new entities as for the previously integrated entities, e.g. by using some attribute-based blocking key such as birth year for persons or manufacturer for products.  Several works~\cite{christophides2020bigdataER, papadakis2020blocking} have proposed further improvements over such a base approach for specific cases and blocking approaches. One approach~\cite{10.1145/2661829.2661869,dynamicBlk,karapiperis2018edbt} is to keep the blocking keys in a data structure with efficient data access to make comparisons among entities faster. Another method to speed up the computation is the use of so-called summarization techniques~\cite{karapiperis2018edbt}. 
One such approach summarizes (divides) larger blocks  into multiple sub-blocks with a representative and directing a new record (query) to the sub-block with the most similar representative. This enables  a constant number of comparisons for each new record which is valuable for both incremental and streaming ER.  
While most blocking approaches rely on domain or schema knowledge, there are also so-called schema-agnostic blocking schemes for highly heterogeneous data where entities of a certain type can have different sets of attributes. Hence, schema-agnostic approaches consider most or all  attribute values and their components (e.g. words or tokens), regardless of the associated attribute names. While there are many schema-agnostic blocking approaches for non-incremental ER~\cite{christophides2020bigdataER, papadakis2020blocking} schema-agnostic blocking approaches for incremental or streaming ER have only recently been proposed~\cite{brasileiro2019incremental,araujo2020acm,gazzarri2021end}.

The \textit{matching} step of incremental ER is limited to the new entities and involves a pair-wise comparison with the existing KG entities determined by the preceding incremental blocking step. The main goal is to determine all similar entities as potential match candidates as input for the final clustering step, where it is decided whether the new entity is added to an existing cluster or whether it should form a new cluster. Pairwise matching can be done as for batch-like ER and is based on the combined similarity between two entities derived from property values or related entities. The match approach can be configured manually, e.g., together with some similarity threshold that should be exceeded for match candidates, or by applying a supervised machine learning model~\cite{christophides2020bigdataER}. If the pairwise match relationships between previously integrated KG entities are maintained in a similarity graph spawning the previous clusters, this graph can be extended by the new entities and links to the newly determined mach candidates as an input for incremental clustering~\cite{saeedi2018csimq}.  

While there are many approaches for batch-like \textit{entity clustering}~\cite{hassanzadeh2009framework,saeedi2017adbis}, the incremental maintenance of entity clusters for new entities has received comparatively little attention. 
A straight-forward approach is to simply add a new entity either to the most similar existing cluster or to create a new cluster if there is no previous cluster with a high enough similarity exceeding some predefined similarity threshold~\cite{welch2012}. However, this approach typically suffers from a strong dependency on the order in which new entities are added. In particular, wrong cluster decisions, e.g., due to data quality problems, will not be corrected and can lead to further errors when new entities are added. 
A more sophisticated incremental approach based on correlation clustering is proposed in  
\cite{gruenheid2014vldb} that maintains previous clusters within a similarity graph. The updated similarity graph is not only used to determine the clusters for new entities but to also repair previous clusters, e.g. by splitting and merging clusters or by  moving entities among clusters. 
The incremental approaches in~\cite{nentwig2018incremental,saeedi2020incremental} support optimized clustering decisions for duplicate-free (sometimes called clean) data sources from which at most one entity can participate per cluster of matching entities. In this case, an effective clustering strategy is the so-called "max-both" approach where an entity $s$ from a set of new entities is only then added to the most similar cluster $c$ when there is no other new entity that is more similar to $c$ than $s$. The approach of~\cite{saeedi2020incremental} also supports a light-weight cluster repair called n-depth reclustering where only entities close to new entities in the updated similarity graph are considered for changing clusters. 

\subsubsection{Entity Fusion}
\label{sec:fusion}
Merging multiple records of the same real-world entity into a single, consistent, and clean representation is referred to as data fusion~\cite{bleiholder2009data}.
This is a main step in data integration 
as it combines information from several entities into one enriched entity.
Data fusion still entails resolving inconsistencies in the data. First the records may disagree on the names of matching attributes so that one preferred name has to be chosen that should be the consistent with the attribute names of other entities of the same type to facilitate querying. Furthermore, the matching records can disagree on the values of an attribute. There are three main strategies to handle such attribute-level inconsistencies or conflicts~\cite{bleiholder2009data}:

\begin{itemize}
\item Conflict Ignorance: The conflict is not handled but the different attribute values may be retained or the problem can be delegated to the user application.
\item Conflict Avoidance: It applies a unique strategy for all data. For example, it prioritizes data from trusted sources over others.
\item Conflict Resolution: It considers all data and metadata before applying a decision to apply a specified strategy, such as taking the most frequent, most recent or a randomly selected value.
\end{itemize}

Such techniques were first applied for relational data but also found use for Linked Data fusion~\cite{bizer2012ldif}. 
A valuable strategy is to combine multiple value scoring functions.
Mendes et al.~\cite{mendes2012sieve} combine two methods named \textit{TrustYourFriends} (prioritizing data from the trusted source) and \textit{KeepUpToDate} (using the latest value) for conflict avoidance and resolution. 
Moreover, they apply input quality assessment metrics to filter out the values below a threshold or keep the values with the highest quality assessment. Other techniques such as computing average, minimum, and maximum or taking the most frequent values are provided by their data integration framework. 
Dong et al.~\cite{Dong2013DataFR} combine \textit{TrustYourFriends} with a \textit{Weighted Voting} (most frequent or similar values) approach, whereas the former source ranking score is calculated based on an statistical approach.
Similarly, Frey et al.~\cite{frey2019dbpedia} apply a median-based approach for Linked Data fusion. They distinguish functional properties\footnote{A functional property is specified to have a maximum cardinality of one. e.g., a person entity should only have one value for date of birth.} and assign a single value to them. Non-functional properties can be assigned multiple different values.

\subsection{Quality Assurance}
\label{sec:evaluation}

The quality of a KG is crucial for its credibility and therefore its usability in applications~\cite{10.1080/07421222.1996.11518099}.
Quality assurance is the task of maintaining a high KG quality despite the continuous evolution of the KG.
It comprises \textit{quality evaluation} to assess the quality and detect quality issues
as well as \textit{quality improvement} to fix or mitigate quality issues by refining, repairing, and completing the KG. We discuss \textit{knowledge completion} in the next subsection \ref{sec:completion},
due to its unique nature of adding data to the KG  rather than improving or removing existing information.
Quality assurance is important not only for the resulting KG as an outcome of the KG construction process but also within the different construction tasks, such as selecting good-quality sources (Section~\ref{subsubsec:DataAqc}), data cleaning for acquired data, knowledge extraction, ontology evolution or entity fusion. 
The data cleaning approaches mentioned in Section \ref{sec:datacleaning} can also be applied to the KG, e.g., to identify outliers or contradicting information.
Metadata such as provenance information is also important for quality assurance, for example, to explain and maintain KG data concerning the context and validity of conflicting values~\cite{Weikum2021MachineKC}.

Assessing the quality of a KG is extremely challenging since there are many valid ways to structure and populate KGs and even subproblems such as evaluating the quality of a KG ontology are already difficult ~\cite{tartir2005ontoqa,mcdaniel2019evaluating}. In general, evaluating the quality depends on the scope of a KG and should be easier for domain-specific KGs than for very large KGs covering many domains for which completeness may not be possible.  Moreover, the intended KG use cases influence the quality needs of the KG and should thus be considered for KG construction and KG evaluation.   
For example, e-commerce KGs such as  the Amazon Product Graph~\cite{Dong2020AutoKnowSK} should provide up-to-date and reliable product information asking for the enforcement of quality criteria such as accuracy and timeliness. 
Applications such as financial risk analysis also need accurate and trustworthy information from KGs such as the Bloomberg Knowledge Graph~\cite{ridho_reinanda_2021_4903274}. On the other hand, there may also be use cases for which approximate answers - and thus reduced KG quality - may be sufficient, e.g., for obtaining recommendations (similar products, related literature) or to receive aggregated information (e.g., about the relative average income in different countries).

We begin our overview of quality assurance by identifying and describing important quality dimensions. Then, we explore various evaluation methods to measure these dimensions. Next, we investigate correction methods to improve and rectify quality issues. Finally, we present quality evaluation frameworks and benchmark datasets that facilitate quality assessment.

\subsubsection{Quality Dimensions}

KG evaluation  typically involves analyzing various dimensions of quality and the relevance of the dimensions typically depends on the intended kinds of KG usage.
Quality dimensions can be correlated and possibly impact each other positively or negatively, e.g., completeness can negatively affect accuracy. 
Wang et al. identified and surveyed in  ~\cite{Wang2021KnowledgeGQ} six main quality dimensions (accuracy, consistency, timeliness, completeness, trustworthiness, availability) for use in KG evaluation:   

\begin{itemize}
    \item Accuracy indicates the correctness of facts in a KG, including type, value, and relation correctness. It can be separated into syntactic accuracy, assessing wrong value datatype/format, or semantic accuracy, assessing wrong information.
    
    \item Consistency ensures coherency and uniformity of the data within the graph. A consistent KG follows logical rules, avoids contradictions, and maintains coherence among entities, relationships, and attributes. Inconsistencies arise from conflicting information, duplicates, or rule violations.
    
    \item Timeliness in the context of KGs refers to the currency and freshness of the information present in the graph. KG timeliness  can be influenced by the chosen integration approach which may involve batch processing at specific intervals or real-time updates.
    
    \item Completeness captures and reflects knowledge coverage within a specific domain.
    Completeness is also a goal for KG completion  as it involves generating new values or data to augment the current KG. 
    %Unlike other dimensions that may require removing or correcting existing values, KG completion focuses on adding entirely new missing values. 

    \item Trustworthiness indicates confidence and reliability of the KG and depends on source selection and the applied construction methods.
    It is strongly related to the quality dimensions of completeness, accuracy, and timeliness.

    \item Availability is the extent to which knowledge is convenient to use. In other words, it refers to how easily and quickly the knowledge of KGs can be retrieved concerning query complexity and data representation.
    % By addressing the gaps in the knowledge graph, completeness contributes to a more comprehensive representation of the domain or topic at hand.
\end{itemize}

\subsubsection{Evaluation Methods}

A common approach involves crowdsourcing techniques or expert knowledge in the evaluation process of knowledge graphs. During the validation phase, curators can spot errors or verify facts.
Using an iterative human-in-the-loop process allows for continuous improvement and refinement, enhancing the overall reliability and trustworthiness of the graph's data.
One conventional approach is to evaluate the accuracy of the KG against a manually labeled subset of entities and relations~\cite{paulheim2017knowledge}. However, this is costly, so those manually labeled gold standards are usually small. Other approaches use statistical methods such as distance-based, deviations-based, and distribution-based methods~\cite{bizer2009quality}. 
Acosta et al.~\cite{acosta2013crowdsourcing} leverages the wisdom of the crowds in two ways. They launched a contest targeting an expert crowd in order to find and classify erroneous RDF triples and then published the outcome of this contest as paid microtasks on Amazon Mechanical Turk (MTurk) in order to verify the issues spotted by the experts. Their empirical evaluation on DBpedia shows that the two styles of crowdsourcing are complementary and that crowdsourcing-enabled quality assessment is a promising and affordable way to enhance data quality. 
Paulheim et al.~\cite{paulheim2017knowledge} define \textit{retrospective evaluation} as a method in which the human judges the correctness of the KG. The reported quality metric is accuracy or precision. Since KGs are often voluminous, the retrospective approach is typically restricted to a KG sample. 

Another method to evaluate quality is statistical analysis, identifying outliers, inconsistencies, or abnormal data distributions based on patterns and the data structure, including clustering, correlation analysis, or anomaly detection techniques~\cite{Senaratne2021UnsupervisedAD}.

Further, semantic reasoning and inference allow for the validation of the KG's consistency based on the given ontology or individual structural constraints. One method is to calculate disjoint axioms by identifying wrong type statements based on existing relations (e.g., domain and range check)~\cite{Ma2014LearningDA}.
  
Data profiling and cleaning techniques can be applied to find erroneous values based on their distribution. Duplicate detection, schema matching, or entity resolution can be used to identify and resolve inconsistencies, redundancies, or errors (format errors).

Another way of quality evaluation relies on aligning and comparing the entities of the KG with external knowledge and reference sources.
Li et al.~\cite{li2017knowledge} investigate the correctness of a fact by searching for pieces of evidence in other knowledge bases, web data, and searching logs. 
Similarly, \cite{lehmann2012defacto} suggests individual checking of a single fact in different datasets in order to detect inaccurate facts.
This is also useful for yielding larger-scale gold standards but has two sources of errors: errors in the target knowledge graph and errors in the linkage between the two.

Finally, a rule-based analysis is a common solution to detect quality issues based on manually generated constraints, e.g., value restriction or allowed/wanted properties for a specifically typed entity~\cite{shacl,Bonifati2022PGSchemaSF}. 

\subsubsection{Quality Improvement}
\label{sec:qa-improvemnt}

Quality improvement aims at optimizing the KG, making it more reliable, useful, and valuable for its intended purpose and domain.
Quality improvement activities include and combine several task areas discussed in the former sections.
Data cleaning (Section~\ref{sec:datacleaning}) addresses errors, inconsistencies, and redundancies in the graph. Error correction techniques eliminate incorrect or outdated information and adjust inconsistent data entries. Outlier detection identifies and handles data points deviating significantly from the norm.
Entity resolution (Section~\ref{sec:entity-resolution}) methods merge or link entities referring to the same real-world entity. Data fusion(Section~\ref{sec:fusion}) integrates information from multiple sources to enhance overall data quality. Continuous ontology development (Section~\ref{sec:ontology-development}) refines and expands the graph's underlying ontology to accommodate new knowledge and evolving requirements.

Instead of filling in missing data, it may be preferable to remove irrelevant entities that do not pertain to the intended domain. This will prevent the KG from being unnecessarily bloated. Applying automatic approaches can cause extraction of irrelevant information and requires techniques either of manual nature or by leveraging known information from external already structured databases.

In KG, quality assurance, versioning, and rollback mechanisms are crucial for managing errors and maintaining data integrity. By implementing version control mechanisms, changes in the KG can be tracked, allowing for easy rollback in the event of errors or quality issues. This ensures that previous versions of the KG can be restored, providing a safety net for data consistency. Furthermore, maintaining an audit trail of changes and ensuring traceability supports data governance and reproducibility. Section~\ref{sec:metadata-management} discusses various approaches to versioning that can be applied in this context.

\subsubsection{Frameworks and Benchmarks}

The importance and complexity of KG quality assessment and improvement asks for powerful frameworks and tools to support these tasks. A quality evaluation framework incorporates metrics and processes to evaluate quality dimensions, ensuring a clear understanding of the graph's quality aligned with specific applications or use cases.
Further, such a framework can already support mechanisms and techniques for quality improvement, either requiring a human-in-the-loop approach or applying automatic error correction.

Chen et al.~\cite{Chen2019APF} give an overview of the requirements of KG evaluation frameworks, focusing on specific domains.
A special requirement is the scalability of such a framework to be applicable to a huge amount of data.
Considering the degree of automation, using human-in-the-loop approaches might require KG sampling to only evaluate sub-graphs of the entire KG.

There exist already several frameworks and tools for KG quality evaluation and benchmarking. 
TripleCheckMate~\cite{Kontokostas2013TripleCheckMateAT} is a crowdsourcing tool that allows users to evaluate single resources in a RDF KG, by annotating found errors with one of 17 error classes.
Another human-in-the-loop approach was proposed by NELL where learned extraction patterns were validated by a user after a certain number of iterations~\cite{carlson2010toward/nell}.
RDFUnit~\cite{Kontokostas2014TestdrivenEO} is an evaluation tool for validating and testing RDF graphs against predefined quality constraints and patterns. It can assess the quality and compliance of RDF datasets concerning schema definitions, vocabulary usage, and data integrity (supporting SHACL).
The tool enables the automatic generation of tests by analyzing the structure of schemata, like ontologies or vocabularies, and generating test cases based on defined rules or patterns.
Additionally, the tool allows users to define custom validation rules or include existing vocabularies and ontologies for validation purposes. 

Hobbit~\cite{HOBBIT} (Holistic Benchmarking of Big Linked Data) is a platform that facilitates benchmarking of linked data systems and components. It provides a standardized framework for evaluating and comparing algorithms and approaches used in processing linked datasets. Key features include configuring benchmarking workflows, evaluating performance metrics, visualizing results, and supporting reproducibility and sharing of benchmarks. 

Benchmark datasets exist for specific subtasks of KG construction, such as entity resolution (e.g. Gollum~\cite{hertling2022gollum}) and knowledge completion (e.g. CoDEx~\cite{Safavi2020CoDExAC}).
While there are several benchmark datasets available that focus on specific subtasks of the construction process, there is a lack of widely used end-to-end benchmark datasets, and researchers often create custom datasets or use subsets of existing datasets to evaluate their construction pipeline.

\subsection{Knowledge Completion}
\label{sec:completion}
Knowledge Graph completion is the task of adding new entries (nodes, relations, properties) to the graph using existing relations.
Paulheim~\cite{paulheim2017knowledge} surveys KG completion approaches as well as evaluation methods. He also distinguishes internal from external methods, especially for determining missing entity type information and relations. Internal approaches solely rely on the KG as input, whereas external methods incorporate additional data like text corpora, and in a broader context, human knowledge sources like crowdsourcing. The survey concludes that current approaches for KG completion typically limit themselves to a single task such as determining missing type information, or missing relations (link prediction) or missing attribute values (literals). 
Holistic solutions to simultaneously improve the quality of KGs in several areas are thus currently missing.

\subsubsection{Type completion}
Type completion refers to the task of assigning types to nodes without type information.
In the case of PGM it is in most cases not allowed to have nodes without type information~\cite{Angles2018ThePG}. Since there is limited standardization in the realm of PGM~\cite{DBLP:conf/amw/AnglesTT19}, and the possibility to label nodes with unknown type with a dummy label, type completion can still be seen as a relevant KG completion task. 
In this case node classification approaches can be used to predict classes of unlabeled nodes. For example Neo4j provides a specific node classification pipeline\footnote{\url{https://neo4j.com/docs/graph-data-science/current/machine-learning/node-property-prediction/nodeclassification-pipelines/node-classification/}}, although the resulting predictions are added as node properties necessitating a post-processing step to redefine the label.

The traditional way of determining missing type information in RDF datasets involves the use of logical reasoning.
However this approach is limited since it relies on already consistent facts and existing \texttt{rdf:type} information in the knowledge base~\cite{DBLP:conf/semweb/PaulheimB13}.
To address this shortcoming statistical approaches use the distribution of relations between entities to predict missing type information.
For example SDType~\cite{DBLP:journals/ijswis/PaulheimB14} uses a weighted voting approach based on the statistical distribution of the subject and object types of properties.
Similarly StaTIX~\cite{DBLP:conf/bigdataconf/LutovRKC18} relies on weighted statistics of multiple properties of entities as input for their clustering approach.

Recently, there has been some attention on leveraging KG embeddings to infer type information. For example ConnectE~\cite{zhao-etal-2020-connecting} incorporates two mechanisms with one relying on \textit{local typing knowledge} and the other on \textit{global triple knowledge}. 
The first relies on the fact that entities close in the embedding probably share the same type.
Relying on relationship information the second mechanism learns entity type embeddings by replacing the subject and object entity in a triple for their corresponding type.
Finally, for entity type prediction a composite score of the two mechanisms is used.

\subsubsection{Link Prediction}
The task of link prediction aims at finding missing relations in a KG.
In the example presented in Figure~\ref{fig:kgdef} we see that while there is a \textit{writtenBy} relation between the song \textit{Xtal} and the artist \textit{Aphex Twin}, there is no relation between the song \textit{Ageispolis} and \textit{Aphex Twin}. Predicting this missing \textit{writtenBy} relation would be a goal of link prediction.

Based on~\cite{paulheim2017knowledge}, a common method for the prediction of a relation between two entities is distant supervision~\cite{aprosio2013extending,gerber2013real,gerber2011bootstrapping,mintz2009distant} using external resources.
This method starts with linking entities of the Knowledge Graph to the text corpus using NLP approaches and then tries finding patterns in the text between entities. Another approach~\cite{west2014knowledge} uses the same methodology, but considers the whole Web as the corpus.
Lange et al.~\cite{lange2010extracting} learn patterns on Wikipedia abstracts using Conditional Random Fields~\cite{fields2001probabilistic}.
Blevins et al.~\cite{blevins2020moving} propose a similar approach, but on entire Wikipedia articles.
Another line of research uses semi-structured data such as tables~\cite{munoz2013triplifying,ritze2015matching} or list pages~\cite{paulheim2013extending} in Wikidata for predicting missing relations.

In the last years considerable research attention was devoted to investigating KG embeddings for the task of link prediction.
These methods encode entities and relations of a KG as low-dimensional vectors in an embedding space.
The existing triples in a Knowledge Graph can be used to train such models, evaluating their performance on a held-out set of triples.
For example TransE~\cite{TransE} encodes relations as translations from subject to object entity of a triple $(subject, predicate, object)$.\footnote{In link prediction literature triples are usually signified as $(head, relation, tail)$. In favor of a consistent nomenclature we use the triple signifiers commonly used for RDF.}
This is done by minimizing the distance between \texttt{s} + \texttt{p} and \texttt{o}, where \texttt{s}, \texttt{p} and \texttt{o} are the embeddings of $subject,predicate$ and $object$ respectively. 
A variety of approaches have been devised to address the problems of TransE to model $1-n$ or $n-n$ relations, by e.g. encoding relations in a separate hyperplane~\cite{TransH} or operating in the hyperbolic space~\cite{HyperKG}.
For a more broad overview and benchmark study we refer to this paper~\cite{PykeenEval}.

The described embedding-based link prediction methods rely on \textit{shallow embeddings}, which means all embeddings are stored in a entity/relation-matrix and obtaining the respective embedding for an entity or relation is done by using a lookup-table.
These approaches are unable to deal with unseen entities. 
The study of \textit{inductive} link prediction aims to address this shortcoming.
GraIL~\cite{GraIL} relies on Graph Neural Networks (GNN) to achieve this.
This approach samples the subgraph enclosing the link to be predicted, then labels the nodes in this subgraph based on the distance to the target nodes (i.e. the nodes which the link would connect). The labeled subgraph is then used in a GNN to score the likelihood of a triple.
NodePiece~\cite{galkin2022nodepiece} can perform inductive link prediction via a compositional representation for entities. Relations around a node are sampled in order to create a node hash, which is passed through an encoder to obtain the final entity embedding. Being able to create entity representation for unseen entities, but known relations permits NodePiece to then use any scoring function (e.g. TransE) for the link prediction task.

A special type of link prediction aims to discover identity links (e.g. \texttt{owl:sameAs} relations)~\cite{hogan2021knowledge}, which connect nodes, that refer to the same entity. This task serves the same goal as entity resolution (discussed in Section~\ref{sec:entity-resolution}).

\subsubsection{Data Enrichment} % Refinement: Completion; Correction; 
Concerning aspects of domain coverage and succinctness, additional processes are applicable that increase the final quality of the KG. 
In addition to type and relation prediction, domain knowledge could possibly be extended by loading completing entity information from external (open accessible) knowledge bases. 
This approach is different to the process of integrating an entire external data collection but only focuses on loading necessary domain information that relates to the already integrated entities.
For enhancing KG data with additional relevant domain entities information external knowledge bases can be requested based on extracted (global) persistent identifiers (PID). For example extracted ISBN numbers, DOIs, or ORCIDs allow to request additional external information from Wikidata; or Gene and Protein data is accessible based on their symbols in public biochemical databases, like the National Library of Medicine\footnote{\url{https://www.ncbi.nlm.nih.gov/}}.
Paulheim surveys approaches that exploit links to other KGs in order to not only verify information but also to find additional information to fill existing gaps~\cite{paulheim2017knowledge}.

\newcommand\ok[0]{
    \tikz[baseline=(char.base)]{
        \node[shape=circle,draw,inner sep=0pt] (char) {\ding{51}};
    }
}
\newcommand\sm[0]{
    \tikz[baseline=(char.base)]{
        \node[shape=circle,draw,inner sep=1.75pt] (char) {\vphantom{?}};
    }
}
\newcommand\uc[0]{
    \tikz[baseline=(char.base)]{
        \node[shape=circle,draw,inner sep=1pt] (char) {?};
    }
}

\newcommand\simple[0]{
	\Circle~
}

\newcommand\complex[0]{
	\CIRCLE~
}

\section{Overview of Knowledge Graph Construction Pipelines and Toolsets}
\label{sec:example-kgs}

\begin{table}[!ht]
\caption{Overview  of selected KGs. '*' in the first column indicates manually curated (crowd-sourced) KGs. '?' means unknown/undisclosed values. The statistics include the KG's year of announcement, targeted domain, processed number of data sources, KG data model,  graph size, number of versions,  and year of last update. Used domain abbreviations: Cross = cross domain, MLang = multi-lingual data.}
\label{tab:kg-overview}
\centering
\addtolength{\tabcolsep}{-2pt}
\begin{tabular}{|ll|c|c|c|c|c|c|c|c|c|c|}

\hline
& & Year & Domain & Srcs. & Model & Entities & Relations & Types & R-Types & Vers. & Update \\

\hline
\multicolumn{2}{|l|}{\underline{Closed  KG}} & & & & & & & & & & \\
& Google KG~\cite{noy2019industry} & 2012 & Cross,MLang & $>>>$1 & Custom,RDF & 1B & >100B & ? & ? & ? & ? \\
& Diffbot.com & 2019 & Cross & $>>>$1 & RDF & 5.9B & >1T & ? & ? & ? & ?\\
& Amazon PG~\cite{Dong2020AutoKnowSK} & 2020 & Products & >1 & Custom & 30M & 1B & 19K & 1K & ? & ? \\
\hline
\multicolumn{2}{|l|}{\underline{Open Access KG}} & & & & & & & & & & \\
& *Freebase~\cite{bollacker2007freebase} & 2007 & Cross & $>>$1 & RDF & 22M & 3.2B & 53K & 70K & $>$1 & 2016 \\
& DBpedia~\cite{hofer2020new} & 2007 & Cross,MLang & 140 & RDF & 50M & 21B & 1.3K & 55K & $>$20 & 2023 \\
& YAGO~\cite{Suchanek2007YagoAC,pellissier2020yago} & 2007 & Cross & 2-3 & RDF(-Star) & 67M & 2B & 10K & 157 & 5 & 2020 \\
& NELL~\cite{carlson2010toward/nell} & 2010 & Cross & $\ge$1 & Custom,RDF & 2M & 2.8M & 1.2K & 834 & $>$1100 & 2018 \\
& *Wikidata~\cite{vrandevcic2014wikidata} & 2012 & Cross,MLang & $>>>$1 & Custom,RDF & 100M & 14B & 300K & 10.3K & $>$100 & 2023 \\
& DBpedia-EN Live~\cite{morsey2012dbpedia} & 2012 & Cross & 1 & RDF & 7.6M & 1.1B & 800 & 1.3K & $>>>$1 & 2023 \\
& Artist-KG~\cite{gawriljuk2016scalable} & 2016 & Artists & 4 & Custom & 161K & 15M & $>$1 & 18 & 1 & 2016 \\
& *ORKG~\cite{auer2020improving/orkg} & 2019 & Research & $>>$1 & RDF & 130K & 870K & 1.3K & 6.3K & $>$1 & 2023 \\
& AI-KG~\cite{dessi2020ai/ai-kg} & 2020 & AI Science & 3 & RDF & 820K & 1.2M & 5 & 27 & 2 & 2020 \\
& CovidGraph~\cite{Covidgraph} & 2020 & COVID-19 & 17 & PGM & 36M & 59M & 128 & 171 & $>$1 & 2020 \\
& DRKG~\cite{drkg2020} & 2020 & BioMedicine & $>$7 & CSV & 97K & 5.8M & 17 & 107 & 1 & 2020 \\
& VisualSem~\cite{Alberts2020VisualSemAH} & 2020 & Cross,MLang & 2 & Custom & 90k & 1.5M & (49K) & 13 & 2 & 2020 \\
& WorldKG~\cite{Dsouza2021WorldKGAW} & 2021 & Geographic & 1 & RDF & 113M & 829M & 1176 & 1820 & 1 & 2021 \\
\hline
\end{tabular}
\end{table}
\begin{adjustbox}{center,caption={Comparison of KG construction approaches w.r.t.\textit{Consumed Data}, generated \textit{Metadata}, and \textit{Performed Construction Tasks}. The construction tasks are rated as \textit{simple/manual} \simple or \textit{sophisticated/semi-automatic} \complex. '?' indicates \textit{mentioned but unclear} implementation. Each criterion can cover multiple solutions.},float=table}
\footnotesize

\label{tab:benchmarkDimensions}
\centering
\begin{tabular}{|ll|c|c|c||ccccc||ccc||cccccccc|}

\hline

\multirow{2}{*}{ } & \multirow{2}{*}{ } & \multirow{2}{*}{} & \multirow{2}{*}{} & & \multicolumn{5}{c||}{Consumed Data} & \multicolumn{3}{c||}{(Meta)Data} & \multicolumn{8}{c|}{Performed Construction Tasks} \\

\multicolumn{2}{|l|}{Name of System} & \rotatebox{90}{System Version/Year} &  \rotatebox{90}{Open Implementation} & \rotatebox{90}{Incremental Integration} & \rotatebox{90}{Unstructured Data} & \rotatebox{90}{Semi-Structured Data} & \rotatebox{90}{Structured Data} & \rotatebox{90}{(Event-)Stream Data} & \rotatebox{90}{Supplementary Input} & \rotatebox{90}{Deep Provenance} & \rotatebox{90}{Temporal Data} & \rotatebox{90}{Additional Metadata} & \rotatebox{90}{KG Initialization} & \rotatebox{90}{Input Cleaning} & \rotatebox{90}{Ontology Management} & \rotatebox{90}{Knowledge Extraction} & \rotatebox{90}{Entity Resolution} & \rotatebox{90}{Entity/Value Fusion} & \rotatebox{90}{Quality Assurance} & \rotatebox{90}{Knowledge Completion} \\ 

\hline
\multicolumn{2}{|l|}{\underline{Dataset Specific}} & & & & & & & & & & & & & & & & & & & \\
& DBpedia & 2019 & \checkmark & & & \checkmark & & & \checkmark & \checkmark & \checkmark & \checkmark & \simple & \complex & \simple & \simple & & & \complex & \simple \\
& YAGO4 & 2020 & \checkmark & & & \checkmark & \checkmark & & \checkmark & \checkmark & \checkmark & & \simple & \simple & \complex & & & & \complex & \\
& DBpedia-Live & 2012 & \checkmark & \simple & & \checkmark & & \checkmark & \checkmark & \checkmark & \checkmark & & \simple & \complex & \simple & \simple & & & & \\
& NELL& 2018 & & \complex & \checkmark & \checkmark & & & \checkmark & \checkmark & \checkmark & & \simple & & \complex & \complex & & & \simple & \\
& Artist-KG & 2016  & \checkmark & \simple & & \checkmark & \checkmark & & & & & & \simple & \simple & \complex & & \complex & & & \\
& AI-KG& 2020 & & ? & & \checkmark & & & \checkmark & \checkmark & \checkmark & & \simple & & & \complex &  & & \simple & \\
& CovidGraph& 2020 & \checkmark & \simple & \checkmark & \checkmark & \checkmark & & \checkmark & \checkmark & ? & & \simple & & ? & \complex & \simple & & & \\
& DRKG & 2020 & \checkmark & & & \checkmark & \checkmark & & & \checkmark & & & ? & & & \simple & & \simple & & \complex \\
& VisualSem& 2020 & \checkmark &  & \checkmark & \checkmark & \checkmark & & \checkmark & & & & \simple & \complex & & \simple & & & & \\
& WorldKG & 2021 & \checkmark & & & \checkmark & & & \checkmark & & & & \complex & \complex & \complex & \simple & & & \simple & \\

\hline
\multicolumn{2}{|l|}{\underline{Toolset/Strategy}} & & & & & & & & & & & & & & & & & & & \\
& FlexiFusion~\cite{frey2019dbpedia} & 2019 & & & & \checkmark & \checkmark & & \checkmark & \checkmark & & \checkmark & \simple & \simple & & & & \complex & & \\
& dstlr~\cite{dstlr} & 2019 & \checkmark & ? & \checkmark & & & & & \checkmark & & & \simple & & & \complex & & & \simple & \simple \\
& XI~\cite{cudre2020leveraging/xi} & 2020 & & ? & \checkmark & \checkmark & & & & ? & ? & ? & \simple & & & \complex & & & ? & \\
& AutoKnow~\cite{Dong2020AutoKnowSK} & 2020 & & & \checkmark & \checkmark & & & \checkmark & & & & \simple & \complex & \complex & \complex & & & & \complex \\
& HKGB~\cite{Zhang2020HKGBAI} & 2020 & & \simple & & \checkmark & & & \checkmark & & \checkmark & & \complex & & \complex & \complex & ? & & \simple & \complex \\
& SLOGERT~\cite{Ekelhart2021TheSF} & 2021 & \checkmark &  & & \checkmark & & & & \checkmark & \checkmark & & \simple & & & \complex & ? & & & \simple \\
& SAGA~\cite{ilyas2022saga} & 2022 & & \complex & \checkmark & \checkmark & \checkmark & \checkmark & \checkmark & \checkmark & \checkmark & & ? & \complex & \simple & \complex & \complex & \complex & \complex & \complex \\ 
\hline

\end{tabular} 

\end{adjustbox}

We now investigate and compare construction pipelines for existing KGs and for KG construction toolsets with respect to the KG requirements and construction steps introduced in the previous sections. The KG-specific approaches focus on integrating data from a rather fixed set of data sources for a single KG while the toolsets (or strategies) are more generic and can be applied for different sources and KGs. 
Overall we consider 16 KG-specific approaches (with a focus on ten semi-automatic and open implementations) and seven toolsets. 
In the first subsection we give an overview of the different approaches including data statistics for the respective KGs and characteristics about the data sources and their construction pipelines. This overview aims already at providing a good assessment of the current state of the art. Section~\ref{sec:kg-comparison} gives a detailed comparison of the approaches w.r.t our requirements.
In the two further subsections we give additional details about the KG-specific approaches and the toolsets. In Section \ref{sec:challenges} we will discuss remaining challenges and thus areas for future work. 

\subsection{Overview}
\label{sec:kg-overview}

Our overview is divided into two parts. We first 
summarize general information and statistics for the selected KGs using Table~\ref{tab:kg-overview} and then investigate KG construction criteria for KG-specific pipelines and KG toolsets using Table~\ref{tab:benchmarkDimensions}. 
Given the enormous and growing number of KGs, we had to restrict ourselves to a small number of efforts. In our selection we try to cover popular KGs such as DBpedia and Yago as well as more current approaches for either a single domain or several domains (cross domain).  
Most importantly, we focus on KG projects described in peer-reviewed articles and discuss \textbf{closed KGs} only briefly as their data is not publicly accessible and the used techniques are not verifiable. 
Such closed KGs are typically developed and used in companies such as company-specific Enterprise KGs~\cite{iceis17} and the KGs of big web and IT companies such as Google~\cite{noy2019industry}, Amazon~\cite{Dong2020AutoKnowSK}, Facebook, Microsoft~\cite{MicrosoftAcademicKG}, Tencent, or IBM. 
However, open and easy to use KG toolsets are still in their infancy. 
Here we tried to include recently described approaches that have already been applied to create several KGs including those for a specific domain or a single data type. 
In order to obtain a representative sample over the state-of-the-art we employed a keyword-based search in academic search engines (e.g. SemanticScholar) and Github to gather all papers and approaches (systems and toolsets) that might fit our criteria. We also relied on existing surveys to gather potentially missed approaches. After a manual selection process via the paper titles and abstracts and comparison with our requirements we created a list of 64  candidate sytems. This methodology is in line with other surveys~\cite{DeclarativeAsscheReview23}. After closer inspection w.r.t. the coverage of our requirements and filtering the approaches by age and availability of documentation or publication we were left with the 23 works that are described in detail here.
We expect that our criteria for comparison and methodology are also useful to evaluate KG-specific and more generic construction approaches not covered in this paper.

Table~\ref{tab:kg-overview} summarizes general characteristics of the selected KGs which are grouped into \textit{closed} and \textit{open access} KGs and in each group ordered by their year of announcement or first publication. The table also displays the KG's targeted domain, processed number of data sources, underlying data model, graph size (number of entities, relations, entity types and relation types), number of versions and year of last update. 
Table~\ref{tab:kg-overview} excludes the toolset projects as these are not restricted to a single KG.
The values in this table were obtained from the most recent available version, either from the publication or directly from the dataset. The date of this version is denoted in the "Update" column in the table.

The table includes three manually curated projects based on crowdsourcing, namely the well-known Freebase and Wikidata approaches as well as the newer Open Research Knowledge Graph (ORKG). 
Freebase~\cite{bollacker2007freebase,pellissier2016freebase} was one of the first collaboratively built and versioned KGs and, after its shutdown in 2016, became a popular source for building several other KGs, like  Wikidata~\cite{vrandevcic2014wikidata}.
Wikidata supports entity annotation with key-value pairs, including validity time, provenance, and other metadata, such as references~\cite{piscopo2017provenance}. It facilitates semi-automatic curation involving both bots and human curators.
As a project of the Wikimedia Foundation, full data dump snapshots are released twice a month.
The ORKG~\cite{auer2020improving/orkg} focuses on publications where manually uploaded papers are automatically enriched with metadata.
The platform provides tools to extract information such as tables and figures from publications and to help find and compare similar publications of interest. 

Most of the considered KGs are based on RDF while some use a property graph or custom graph data model (fifth column in Table~\ref{tab:kg-overview}).
Regarding the covered domains, the selected KGs either integrate sources from different domains (cross-domain) or focus on a single domain such as research, biomedicine or Covid-19.
A possible limitation of cross-domain KGs, especially for smaller-sized ones, is that they can miss domain-specific details or expert knowledge.
Some of the KGs contain and connect multilingual information (MLang) by providing descriptive entity values in different languages. These translations are mostly taken directly from one of the sources (e.g., Wikipedia or BableNet), instead of generating own translation during the construction process.
There are large differences among the KGs regarding the number of integrated source datasets (from 1 to 140) and the size of the KGs in terms of number of entity and relation types and number of entities and relations. With the highest number of sources, DBpedia independently extracts 140 sources (Wikipedias), with equivalent entities being interlinked by the extracted \texttt{sameAs} connections contained in the page articles.
The closed KGs are by far the largest with up to almost 6 billion entities and more than a trillion relations (Diffbot.com). Wikidata is the largest open-source KG with about 100 million entities of 300K entity types and 14 billion relations of 300K relation types. The smallest KGs have less than 1 million entities or relations. In general, the open KGs are rather limited in the number and diversity of the data sources while closed approaches such as Google KG aim at integrating information at web scale. 
Only a few of the KG projects continuously release updated versions of their KG while most projects only released data once or irregularly every few years. This underlines that continuous maintenance of KGs is not yet commonplace. With over 1100 dumps\footnote{\url{http://rtw.ml.cmu.edu/rtw/resources}} NELL features the highest number of continuously and incrementally generated KG versions.

\subsection{Comparison}
\label{sec:kg-comparison}

We now turn to a closer inspection of the KG construction processes of the individual KGs and toolsets. 
Table~\ref{tab:benchmarkDimensions} summarizes the corresponding information for the ten open access KGs with semi-automatic construction as well as for seven toolsets / strategies for KG construction. 
We derived the concrete set of comparison criteria from our previously specified KG requirements in Sec.~\ref{sec:requirements} such as support for incremental updates and different input data. Other criteria relate to the individual construction tasks from Sec.~\ref{sec:tasks} necessary to meet the requirements, e.g. to support certain kinds of input data (e.g., knowledge extraction or entity resolution tasks) or to meet the requirement of quality assurance (tasks of input cleaning, quality assurance, and knowledge completion).  
In Table~\ref{tab:benchmarkDimensions} we weighted the criteria with regards to their fulfilment/presence in a specific solution by indicating strong approaches (considering automation, quality and flexibility) with a full circle and weaker approaches open circle symbol, compared to the others approaches. We will give a detailed explanation on each of the criteria decisions for each of the approaches in Section~\ref{sec:kg-specific-comparison} and~\ref{sec:kg-toolset-comparison}.
We also provide information on the year of the considered version (publication) and indicate whether the approach offers an open implementation. We see that the pipeline/toolset implementations for two of the open access KGs (NELL, AI-KG) and even five of the seven toolsets are closed-source including the approaches from Amazon (AutoKnow) and Apple (SAGA). 

The third column in Table~\ref{tab:benchmarkDimensions} indicates to what degree incremental KG updates are supported, i.e., that changes in the data sources or new sources can be integrated without a full recomputation of the KG. 
We see that most approaches have either no or unknown support for incremental updates. DBpedia and Yago are limited to batch updates with a full recomputation of the KG while others such as DRKG and WorldKG represent one-time efforts without KG updates at all. 
Other approaches provide simple incremental capabilities. Artist-KG and CovidGraph are able to integrate new sources incrementally, but do not specify ways to ingest changes in the underlying data sources.
DBpedia-Live tracks changes automatically in the underlying data sources and integrates them directly whilst skipping expensive quality assurance steps.
Two approaches provide sophisticated incremental capabilities. NELL grows a KG via a semi-supervised approach enabling human interaction to avoid accumulation of errors. SAGA is one of the most sophisticated approaches w.r.t incremental integration. It has the ability to ingest changes into a stable KG, which is updated in batches, and also serves a live KG which forgoes some quality assurance steps in lieu of data freshness.
For some approaches their incremental integration capabilities are unclear. AI-KG describe such capabilities as future work and dstlr mention the ability to track document changes via Apache solr, but do not mention any ways to deal with such changes. In the case of the XI pipeline, this feature's support was considered in the final implementations.

In the following, we describe and discuss the further criteria considered that fall into three groups regarding the consumed input data, generated metadata and the performed construction tasks. 

\textbf{Consumed Input.} For the input data we differentiate the supported kind of data (\textit{unstructured, semi-structured, structured}) and consider support for stream input data 
and supplementary input data for further processing steps (e.g., metadata, mappings, but excluding tool configurations). 

As Table~\ref{tab:benchmarkDimensions} shows, populating KGs from semi-structured data is most common while only about half of the considered solutions or toolset support the import from unstructured or structure data. Several popular KGs (DBpedia, YAGO, NELL) integrate information from Wikipedia and use it as a premium source for a high amount of valuable knowledge. 
Open accessible databases such as WordNet, ImageNet, or BabelNet are also frequent starting points for KG construction. 
Only two of the projects support the continuous consumption of event streams (DBpedia Live and SAGA). NELL continuously crawls the web for new data but updates the KG in a batch-like manner. 
Most approaches integrate supplementary data, especially mapping rules, training data, or quality constraints (SHACL shapes).

\textbf{Collected Metadata.} We consider whether deep or fact-level provenance, temporal information (e.g, validity time) and additional metadata such as aggregated statistics, process reports, or descriptive and administrative information are collected and added to the KG or a separate repository.  

The acquisition of provenance data is the most common kind of metadata support and ranges from simple source identifiers and confidence scores up to the inclusion of the original values. Several systems maintain temporal metadata while further metadata is hardly supported or at least not described.  
In the case of the toolsets, the generation of additional metadata is possible in XI but depends on the use case and resulting pipeline. In general, support for metadata is thus limited and has room for improvement. 

\textbf{Construction Tasks.} In this group we consider to what degree the eight construction tasks introduced in the previous section are supported.   

\begin{itemize}
\item \textbf{KG Initialization} - 
Here a common strategy is to manually create the initial KG either by development from scratch or reusing existing KGs. There may also be a complex pipeline to construct the initial KG by processing semi-structured data from catalogs, wikis, or category systems. All projects start with building or using some initial KG data.
WorldKG and HKGB semi-automatically build an initial ontology and are therefore more advanced compared to a manual ontology construction.
\item \textbf{Input Cleaning (Filtering, Correction)} - support for filtering, normalization, or correction of noisy input data. We exclude here NLP/text pre-processing as this is normally part of knowledge extraction.
This functionality is not always provided (or documented) and often based on manually defined rules and filter definitions, e.g., to select properties and relationships for certain entity types. 
Some solutions also apply normalization steps, e.g., to unify date or number representations.
\item \textbf{Ontology Management} - most approaches have at least some basic (manual) support to evolve the KG ontology and schema data for newly structured input data. In DBpedia, the KG ontology (and data mappings) can be changed manually and needs to be loaded before running a new batch update. The more freshness-oriented approach of DBpedia Live continuously watches ontology changes and immediately schedules affected entities for re-extraction.
More advanced approaches rely on a semi-automatic ontology evolution or enrichment. In particular, some systems can identify new entity and relation types in the input data for addition to the ontology after manual confirmation (NELL, HKGB). 

While for example WorldKG relies on an unsupervised ML approach for ontology alignment, most approaches still perform alignment and merging of ontologies manually.

\item \textbf{Knowledge Extraction} -  many solutions use rule-based mappings to extract entities and relations from semi-structured sources (DBpedia, Yago, DRKG, VisualSem, WorldKG).
Some tools use machine learning approaches for extraction (AI-KG, CovidGraph, dstlr, SLOGERT, NELL). 
For entity linking different approaches are used such as dictionary-based approaches relying on gathered synonyms (e.g. AI-KG), use of human interaction (XI), or applying entity resolution (e.g. HKGB). 
Given the focus on semi-structured data sources, the techniques for knowledge extraction are generally relatively advanced compared to other steps in KG construction. This has also been made possible by the frequent use of existing knowledge extraction tools such as Stanford CoreNLP, as will see in the discussion of the approaches in the next subsections. 
\item \textbf{Entity Resolution} - 
this task is supported in only by few approaches and the pipelines that do employ ER tend to use sophisticated methods like blocking to address scalability issues (ArtistKG, SAGA), and machine-learning-based matchers (SAGA). HKGB's description of their ER solution is too vague to make a definite statement and for SLOGERT it is mentioned, that in some cases ER might be necessary, but should be done with an external tool.
For textual data, identification and matching of entities to KG elements is already covered by entity linking in the knowledge extraction step (Sec.~\ref{sec:knowledge-extraction}). 
\item \textbf{Entity Fusion} - this is the least supported task in the considered solutions. 
None of the dataset specific KGs performs classical (sophisticated) entity fusion in the manner of consolidating possible value candidates and selecting final entity ids or values. 
Instead, the final KG often contains a union of all extracted values, either with or without provenance, leaving final consolidation/selection to the targeted applications.
The DRKG project uses a simple form of entity fusion to normalize entity identifiers. 

Even for the discussed toolsets the coverage of this task is relatively low. The FlexiFusion allows to apply specific fusion functions, leverages provenance information and performs a stable id assignment for entity and property clusters. SAGA refers to the usage of several truth discovery and source reliability-based fusion methods.

\item \textbf{Quality Assurance} - 
human-in-the-loop strategies have been applied to varying degrees, with some solutions, such as HKGB, relying heavily on user interaction. In contrast, others require only final user approval of the correctness of extracted values or patterns, like NELL. The World KG approach manually verifies all matches to the external ontologies.
Further, SAGA tries to detect potential errors or vandalism automatically. It quarantines them for human curation, where changes are treated directly in the live graph and later applied to the stable graph.

Only DBpedia and YAGO perform an automatic consistency check.
Additionally, YAGO guarantees ontological consistency by applying a logical reasoner, and DBpedia checks for dataset completeness and measures quality against the former last version.

In our study, only dstlr offers support for validating extracted facts against an external knowledge base.

\item \textbf{Knowledge Completion} - the integrated KG data is enriched with locally inferred (relations, types) or external knowledge. 
DBpedia attaches additional entity type information based on current ontology and relation data.
Three approaches (DRKG,HKGB,SAGA) presented ML-based link prediction on graph embeddings to find further knowledge. In the case of the DRKG and HKGB approach it is unclear if the newly predicted information flows back into the KG or is stored separately.

Regarding enrichment with external knowledge: While dstlr links entities to Wikidata, it also fetches stored properties from this external source. However, SLOGERT only adds links to external information based on previously extracted identifiers (PIDs).
\end{itemize}

Overall, we see that the KG-specific approaches have a number of limitations regarding scalability to many sources, support for incremental updates and in several steps regarding metadata, ontology management, entity resolution / fusion, and quality assurance. The toolsets are generally better in terms of their functionality but they are mostly closed-source and thus not usable for new KG projects or research investigations.  

\subsection{KG Specific Solutions}
\label{sec:kg-specific-comparison}

\textbf{DBpedia}~\cite{Auer2007DBpediaAN} is one of the most popular KGs, establishing a central access point for the Semantic Web. It extracts structured data from Wikipedia article dumps utilizing the DBpedia Extraction Framework (DIEF), which was forked by several other wiki-based KG projects
~\cite{hofmann2017dbkwik}. 
The DIEF executes numerous extractors, each extracting a specific aspect of the article page, like type information based on the used info-box template or the pages abstract paragraph. One specific extractor, maps semi-structured information from Wikipedia infoboxes\footnote{An infobox is a fixed-format table (usually in the top right-hand corner) to consistently present some unifying aspect that articles share.} to the DBpedia ontology. 
The DBpedia community manually curates DBpedia's ontology and infobox mappings through a publicly accessible mappings mediawiki\footnote{\url{http://mappings.dbpedia.org/api.php}}. The DIEF fetches the latest version of the ontology and mappings from this API endpoint for each extraction run.
In the post-processing phase, a type consistency step checks whether an extracted entity and its relations are violating the is-a or has-a definitions of the DBpedia ontology (e.g., a person entity should not have mechanical doors and hence be a car simultaneously). And a completion phase materializes transitive type (is-a) relations for each entity. 

Since 2020, the current extraction cycle~\cite{hofer2020new} is built around the Databus~\cite{frey2022managing} (meta)data platform, which allows managing data releases, including descriptions, versioning, data quality reports, and automatic metadata generation.
The monthly performed batch extraction consumes data from up to 140 wikis, including several language-specific Wikipedia versions, Wikidata and Wikimedia Commons. Each release is checked for dataset completeness and validated concerning quality.
In the final data, equal entities of the different wikis are connected by \texttt{sameAs} links derived from the \textit{interwiki links} contained in the data sources.

A testing library helps debugging data problems based on SHACL and other integrity tests. 
Deep provenance is supported by linking each extracted value with the revision id of the originating article and the applied extractor.

In addition to DBpedia's dump extraction, \textbf{DBpedia Live}~\cite{Hellmann2009DBpediaLE} is a service that provides a real-time KG by performing continuous extraction of changed Wikipedia contents. 
Changed articles are fetched and reprocessed to extract all relevant values and override (or add/delete) them in the KG. The live extraction also monitors ontology (mapping wiki) changes and schedules all affected pages for re-extraction~\cite{morsey2012dbpedia}.
For improved performance, the live extraction is not applying the mentioned DBpedia post-processing or quality assurance that are thus only performed when the data is completely reprocessed. 

\textbf{YAGO}~\cite{Suchanek2007YagoAC}. The \textit{Yet Another Great Ontology} project initially extracted information about entities of Wikipedia and combined them with an ontology derived from the hierarchically structured WordNet~\cite{Fellbaum-WordNet-1998}. 
In version 2, the KG was extended by temporal information (Wikipedia edit timestamps) and spatial knowledge from GeoNames\footnote{\url{https://www.geonames.org/}}. 
YAGO 3 extends multilingual knowledge by utilizing Wikipedias inter-language links to cover additional values in many other languages.
The latest version \textbf{YAGO 4}~\cite{pellissier2020yago} no longer uses data from Wikipedia (in combination with Wordnet and GeoNames), but collects data from Wikidata~\cite{vrandevcic2014wikidata} and forces it into a taxonomy based on schema.org\footnote{\url{https://schema.org/}} and Bioschemas~\cite{Bioschema}.  
SHACL constraints are used for classes to enforce a strict consistency.  
Manually defined mappings are applied from Wikidata to an initial set of 235 schema.org classes and 116 relations. 
The process iterates over each Wikidata item, filters low-coverage entities, and accepts entities and their types that are transitively connected to one of the initial classes via sub-class relations resulting in the final taxonomy of 10k classes (taxonomy enrichment). 
YAGO maintains fact provenance by using Wikidata's annotations for validity time or external references.

\textbf{NELL}~\cite{carlson2010toward/nell}. The Never-Ending Language Learner is a system 
to incrementally construct a KG from text corpora and web pages.
Per incremental execution (called iteration), it uses NLP-based knowledge extraction to determine entities, their types and relations between entities with the help of patterns.
The central part of learning is the Coupled Pattern Learner (CPL), which memorizes patterns of the form \textit{"X plays for Y"}. In each iteration, the system
learns new patterns and simultaneously applies its previously learned patterns.
Additionally, NELL generates metadata with newly learned patterns and rules. A user manually validates such newly learned patterns regularly (not necessarily for every iteration) before the pipeline uses them as future supplementary input data.
While NELL generates neither RDF nor PGM data,
the Nell2RDF~\cite{GimnezGarca2018NELL2RDFRT} extension transforms its data to RDF and annotates extracted relations with provenance information about the 
used Wikipedia articles.

\textbf{Artist-KG}~\cite{gawriljuk2016scalable}. Gawriljuk et al. create a KG of artists from four different sources that are incrementally added one after the other. 
Per iteration (source), they first filter out artist entities and apply schema mapping with the Karma approach~\cite{KnoblockKarma2012} to map entity properties. Then they perform entity resolution using artist name and birthdate similarities and utilize MinHash/LSH blocking to make this process scalable. 
They only apply the union of matching entities and leave entity fusion (value consolidation) to consuming applications. The final KG is in a custom graph format, serialized as JSON with with nested representations of artist entities. 

\textbf{AI-KG}~\cite{dessi2020ai/ai-kg}. The Artificial Intelligence KG contains over 820K entities derived from over 333k research publications integrating and enriching data from the Microsoft Academic Graph~\cite{MicrosoftAcademicKG}, the Computer Science Ontology~\cite{ComputerScienceOntology}, and Wikidata.
Its internal ontology builds on SKOS\footnote{\url{https://www.w3.org/2004/02/skos/}}, PROV-O\footnote{\url{https://www.w3.org/TR/prov-o/}}, and OWL.
Fact provenance is supported by keeping identifiers to the original papers and information about the applied extraction tool.
The RDF data of the KG is available via a publicly accessible Virtuoso triple store\footnote{\url{http://scholkg.kmi.open.ac.uk/}}.

The pipeline has four components: 1) extractors, 2) entities handler, 3) relations handler, and 4) triple selector.
The extractors use the NLP tools DyGIE++~\cite{WaddenDyGIE2019}, the CSO Classifier~\cite{Salatino2019CSOClassifier}, and Stanford CoreNLP~\cite{StanfordCoreNLP} to extract entities and relations from text for predefined entity types and relations.
To determine possible entities and relations, the system operates an overlapping strategy between the CoreNLP tools like Open Information Extraction, its POS-Tagger, and the topics detected by the CSO Classifier.
Semantically similar entities are clustered (via word embeddings and hierarchical clustering) and manually revised.
The Triple Selector categorizes facts into valid and non-valid triples.
For relations extracted based on patterns between entities derived by the POS-Tagger, an occurrence frequency threshold is used to determine the validity. By contrast, triples extracted by DyGie++ are trusted as this tool has achieved high precision in previous benchmarks.

While the authors planned periodic KG updates, the update process has not been described. 
The AI-KG was subsequently replaced by the CS-KG~\cite{CSKG}.

\textbf{CovidGraph}~\cite{Covidgraph}.
The open domain-specific CovidGraph
integrates over 17 Covid-related sources on publications, authors, genes, proteins, and diseases from publication archives, databases, or open accessible APIs. 
The KG is managed as a Neo4J database. For each data source, 
KG construction uses a modular process where the integration of every source dataset is handled by a separate Docker container dealing with source-specific data cleaning and mapping tasks. 
The final orchestration and order of integration is managed by a separate container\footnote{\url{https://git.connect.dzd-ev.de/dzdtools/motherlode}}. Part of the data is extracted from biomedical paper abstracts via the biomedical language representation model BioBert~\cite{Lee2019BioBERTAP}. The applied matching techniques use string similarities or usable global identifiers like symbols and ids, or for example the Reference Sequence Database~\cite{DBLP:journals/nar/PruittTM07} for genes. The KG contains provenance information about the originating paper, and nodes have modification timestamps. The project provides a schema graph image for exploration purposes but it remains unclear how the underlying ontology (schema) is maintained. 

Due to the global impact of the Covid disease, similar projects emerged like the COVID literature KG~\cite{covid_kg} which uses other integration methods.

\textbf{DRKG}~\cite{drkg2020}. 
The Drug Repurposing KG integrates several biomedical sources (mostly open-access databases) with information about genes, proteins, diseases, and drugs. Its use cases are centered around drug redesign and re-purposing, e.g., utilizing newly discovered relations in the KG.  
Each data source is mapped to a triple structure using crowd-sourced tools like Bio2RDF\footnote{\url{https://github.com/bio2rdf/bio2rdf-scripts}} and rule-based mapping languages. Bio2RDF converts data to a crowd-sourced ontology, thereby providing some initial types and relationships. 
To resolve entities from different sources, the DRKG subsequently tries to map entities of the same type to a common ID space, e.g., the MESH-ID space for diseases.
The final entity IDs contain the originating source thereby supporting backtracking the origin of facts in the KG.
The DRKG project also applies link prediction by utilizing graph embeddings with the TransE model~\cite{TransE}. This was used to predict relations between drugs, genes and the three diseases SARS, MERS, and SARS-COV2 (COVID-19).

\textbf{VisualSem} \cite{Alberts2020VisualSemAH} is a multilingual and multi-modal KG that interlinks images, their descriptions (glosses) and other attributes from Wikipedia articles, WordNet concepts, and ImageNet~\cite{ImageNet} images.
The approach applies a combination of data retrieval, sampling, and cleaning from the ImageNet dataset.
First an initial set of nodes is retrieved from ImageNet\footnote{ImageNet is an image database organized according to the WordNet hierarchy.} for 1000 classes from the "ImageNet Large Scale Visual Recognition Challenge".
Then, in an iterative process additional nodes are collected in the following way: 1) collect neighboring nodes of the current node pool 2) filter images (removing ones not meeting certain quality criteria 3) filter nodes (keep nodes with at least one image and two relation types) 4) update the node pool (accepting only top-k nodes).
In more detail, during the first step relations from the ImageNet graph are mapped to 13 relationship names in the final graph via manually defined mappings.
In the image cleaning step four filters where applied. The system checks for valid image files, removes duplicated images via SHA1 hashing, uses a ResNET-based binary classifier~\cite{Alberts2020ImagiFilterAR} to remove non-photographic images, and leverages OpenAI's CLIP to remove image that do not minimally match any of the node glosses.
The iteration holds after reaching a node pool size of 90k.
The project is accessible in a public git repository\footnote{\url{https://github.com/iacercalixto/visualsem}} and also contains pretrained models for vision and language research.

\textbf{WorldKG}~\cite{Dsouza2021WorldKGAW} integrates the semi-structured data of OpenStreetMap (OSM)\footnote{\url{https://www.openstreetmap.org}} into a geographic KG.
The construction of WorldKG consists of two parts. 
The WorldKG ontology (initial KG) is constructed in the first part based on OSM's "Map" system and OSM wiki data. The Map system allows users to tag nodes, ways, or relations with geographic attributes encoded as key-value pairs, and the OSM wiki data describes these tags.
In a scraping step, all tags that encode geographic class information are fetched from the site, then the obtained key-value pairs are used to infer a class hierarchy. For example, \textit{school=building} resolves to \textit{schoolBuilding} as a sub type of \textit{building}.
All classes from the initial ontology are then aligned (ontology matching) with Wikidata and DBpedia classes using a unsupervised machine learning approach~\cite{Dsouza2021TowardsNS}.
The resulting class alignments are manually verified and cleaned.
In the second part, the construction processes maps OSM data to the final KG structure performing three steps: 1) filter nodes with at least one tag, 2) filter keys and values based on the initial ontology, and 3) create RDF triples.

\subsection{KG Frameworks \& Strategies}
\label{sec:kg-toolset-comparison}

\textbf{DBpedia FlexiFusion}~\cite{frey2019dbpedia} provides a workflow based on the DBpedia Databus platform to fuse Linked Open Data (LOD) datasets into a provenance-rich and uniform knowledge graph. % with a primary focus on fusing
Users have to register the individual datasets (downloadable dumps) as well as entity matches (link sets) across these datasets, either utilizing sets of \texttt{sameAs} relations of the LOD or by first generating such match links by a suitable entity resolution tool. 
Moreover, Non-RDF data has to be transformed to RDF by users. FlexiFusion generates a so-called PreFusion dump per dataset and uses the linked entities to form a cluster using the connected components algorithm.
Matching entities of a cluster are then fused into a single entity for which the property values are selected, e.g. by taking the values from preferred sources. 
The fusion process distinguishes properties to have exactly one
or multiple values depending on the property cardinality in the originating sources.

Two public KGs have been generated with FlexiFusion: 1) Global DBpedia, which integrates and fuses over 175 sources, including the 140 language-specific DBpedia extractions as well as more than 25 other LOD datasets.
2) an effort to create a Dutch National KG\footnote{\url{https://github.com/dbpedia/dnkg-pilot}} integrating eight sources.
For both projects, DBpedia (and its ontology) served as the initial KG and the existing links from the other datasets to DBpedia were used for the clustering and fusion steps. 

\textbf{dstlr}~\cite{dstlr} is a framework for scalable KG construction, relying on Apache Solr\footnote{\url{https://solr.apache.org/}} as document store, an extraction and completion layer build on Apache Spark\footnote{\url{https://spark.apache.org/index.html}} and Neo4j as graph database for the resulting KG. 
The approach utilizes Stanford CoreNLP for all their knowledge extraction steps: Named-entity extraction, relation extraction and linking the extracted mentions to Wikidata. The framework keeps the provenance of those mentions w.r.t to the source documents. The resulting KG is enriched with facts from an external KG by manually defining mappings between CoreNLP relations and Wikidata properties and extracting corresponding facts from this external resource.
The authors mention the possibility of utilizing Cypher queries to verify information in the KG against Wikidata information. While the authors note, that using Apache Solr as document store enables the ingestion of new batches of documents it is unclear how well the pipeline is capable of handling scenarios that can occur in an incremental paradigm, e.g., the deletion or updating of entities.

\textbf{XI Pipeline}~\cite{cudre2020leveraging/xi}. 
This approach focuses on semi-automatically constructing KGs from unstructured or semi-structured documents such as publications and social network content. 
In the initial step, the system applies named-entity recognition on textual input data. 
Probabilistic models and microtask crowdsourcing are combined to accomplish entity linking with a quality outperforming models without a human-in-the-loop paradigm~\cite{ZenCrowd}.
A type ranking step is performed to achieve fine-grained type associations of entities by leveraging the textual surrounding of entities as well as the current KG type hierarchy~\cite{tonon2016contextualized}.
Relation Extraction utilizes distant supervision and an aggregated piecewise convolution network which is trained on existing relations of the KG.

XI is used for several KG projects. ScienceWise~\cite{aberer2011sciencewise} annotates papers with a human-in-the-loop approach to build a research KG. ArmaTweet~\cite{tonon2017armatweet} enables the detection of events, such as natural disasters or terrorist activity, based on anomalies in extracted Twitter tweets.
Guider~\cite{mavlyutov2017dependency/guider} extracts a dependency graph from logs for dependency-driven analytics. The foundation of the KG is a manually developed ontology. 
Guiders metadata tracks multi-level granular provenance over facts, events, sources (posts, users, locations). In some use cases, the XI functionality has been extended, e.g. for metadata support in Guider.

\textbf{AutoKnow}~\cite{Dong2020AutoKnowSK} is a closed-source approach for creating a product KG 
in the domain of retail products at Amazon. 
The system processes existing 
product catalogs and consumer shopping behavior logs leveraging several machine learning approaches and distant supervision for training.
Its architecture consists of a so-called ontology suite and a data suite. 
The ontology suite performs taxonomy enrichment (extraction, attachment) as well as relation discovery.
In the taxonomy enrichment step, new types are extracted from the input product catalogs and customer queries~\cite{Zheng2018OpenTagOA}. 
A GNN approach is applied to place new types into the existing ontology.
Relations are extracted using classification models for attribute applicability and a regression model for attribute importance on product profiles and the user's search, review, or Q\&A data.

The data suite performs data imputation (knowledge extraction), data cleaning, and synonym finding.
Data imputation extracts attribute-value pairs from the product data using a taxonomy-aware tagging approach. %(conditioned on product type). 
It leverages CRF combined with multi-task learning with a shared BiLSTM to simultaneously train sequence tagging and product type categorization.
The data cleaning phase checks extracted attribute-value pairs for correctness based on a transformer-based neural net model.
Synonym finding is based on a supervised approach using a combination of collaborative filtering and a simple logistic regression model.

Most of the training and validation data are derived from the product catalog or customer behavior logs applying distant supervision and, in some cases, utilizing crowdsourcing with Amazon MTurk.
In an experimental execution, AutoKnow backed the construction of a product graph at Amazon with more than 30 million entities and 1 billion relations assigned to 19K entity types and 1K relation types.

\textbf{HKGB}~\cite{Zhang2020HKGBAI}.
The \textit{Health Knowledge Graph Builder} is a platform to semi-automatically construct clinical KGs with heavy human-in-the-loop (HL) involvement.
As input, the system consumes Electronic Medical Records (EMR) consisting of structured and unstructured parts. It semi-automatically processes the data in combination with the clinicians' inputs and produces graph data in OWL and RDF.
The human interaction involves: 1) new concept/relation inspection (approving recommended concepts or relations), 2) adding medical synonym entities based on instances, 3) the annotation of unstructured data based on instances and relations of the current KG, and 4) the definition of mapping rules from EMR to RDF and the extraction of concepts, entities, and relations. The HL inspection mainly contributes to the ontology construction and all other steps to every aspect of the KG (instances and ontology).
Annotations are accepted if there is high agreement across annotators above some confidence threshold (0.81).
Disease-specific information ingestion divides into two phases: 1) building the Concept KG (ontology), and 2) building the Instance KG.
The authors describe an incremental process to add other diseases afterward using a similar strategy (similar to the initial construction).
In addition to the construction tools, the HKGB platform provides three graph tools for data discovery, extraction, and link prediction to support domain-related applications.

The HKGB was utilized to develop the HuadingKG, which consists of about 85 million entities and 265 million relations, initially built with information about cardiovascular diseases and later enriched with information from the Knee Osteoarthritis domain.

\textbf{SLOGERT}~\cite{Ekelhart2021TheSF}. 
\textit{Semantic LOG ExtRaction Templating} is a framework for automated KG construction from log data. The resulting KGs are utilized in security-related applications to detect upcoming threats and vulnerabilities.
It heavily uses known tools, e.g., LOGPAI for log file pattern parsing, Stanford NLP for entity recognition, and the OTTR Engine (LUTRA) with Apache Jena to manage RDF data.
As the KGs foundation, the internal ontology extends a previous vocabulary~\cite{Kurniwan19SemanticsPoster} with mappings to the Common Event Expression\footnote{\url{https://cee.mitre.org/language/}} taxonomy. The implementation is publicly available in a git repository\footnote{\url{https://github.com/sepses/slogert}} and the log ontology is shared online\footnote{\url{https://w3id.org/sepses/ns/log}}.

The workflow runs in two phases.
The first phase extracts data and parameters from log files
and generates RDF templates conforming to the KG ontology. 
For named entities determined by Stanford CoreNLP type and properties are defined from a log vocabulary.
In the second phase, 
the extracted template information is used to convert 
each log file to a graph, and then the log file graphs are integrated into the final KG.
This is done by utilizing appropriate identifiers to connect local context information (e.g., network architecture or organizational structures) 
to key concepts and identifiers of the computer log domain (IP, MAC, URL) and external sources such as vulnerability databases or service inventories. If external knowledge does not align, an entity resolution step is required; the authors mention the SILK framework~\cite{SilkVolzBizer2009} for this task.

\textbf{SAGA}~\cite{ilyas2022saga}.
This closed-source toolset supports multi-source data integration for both batch-like incremental KG construction and continuous KG updates.  
The internal data model extends standard RDF to capture one-hop relationships among entities, provenance (source), and trustworthiness of values.
The system supports source change detection and delta computation using their last snapshots. Based on detected changes, SAGA executes parallel batch jobs to integrate an updated or new source into its target graph.
SAGA's ingestion component requires mappings from new data to the internal KG ontology. This step only requires predicate mappings, as the subject and object fields can remain in their original namespace and are linked later in the process. 
The required mappings are mostly manually defined and stored as supplementary configuration files. Additionally, data can be reprocessed with the HoloClean tool~\cite{HoloClean} for data repair.
SAGA is able to detect and disambiguate entities from text and (semi-)structured sources.
To make the deduplication step scalable it groups entities by type and performs simple blocking to further partition the data into smaller buckets.
A matching model computes similarity scores and machine-learning- or rule-based methods are applicable to determine likely matches.
Correlation clustering~\cite{pan2015correationCls} is then utilized to determine matching entities.
The system tracks same-as links to original source entities to support debugging.
For entity fusion, (conflicting) entity attribute values are harmonized based on truth discovery methods and source reliability to create consistent entities.

In addition to the stable KG (updated in batches), the system can maintain a \textit{Live Graph}, which continuously integrates streaming data and whose entities reference the stable entities of the batch-based KG. 
For scalability and near-real-time query performance, the live graph uses an inverted index and a key-value store. 
SAGA supports live graph curation by using a human-in-the-loop approach. 
The authors mention that SAGA powers question answering, entity summarization, and text annotation (NER) services.

\section{Discussion \& Open Challenges}
\label{sec:challenges}
Our study of existing solutions for KG construction showed that there are many different approaches not only for building specific KGs but also in the current toolsets. This underlines the inherent complexity of the problem and the dependency on different criteria such as the major kinds of input data and the intended use cases. The requirements we posed in Section \ref{sec:requirements} are also not yet met to a larger degree indicating the need for more research and development efforts. 
This is also because there are inherent tradeoffs between the goals of high data quality, scalability and automation~\cite{Weikum2021MachineKC} that ask for compromise solutions. So while it is possible to have a large degree of automation for individual construction tasks, human interaction generally tends to improve the quality significantly. On the other hand, such human interaction can become a limiting factor towards scalability to many sources and high data volume. 

In the following, we discuss open challenges and areas for future work on KG construction that we see. The focus here is on broader issues rather than specific limitations in individual steps. 

\textbf{Incremental KG Construction} We observed that most construction pipelines for specific KGs and in toolsets do not yet support incremental KG updates but are limited to a batch-like re-creation of the entire KG. As already discussed in the requirement section \ref{sec:requirements} this approach has significant limitations and prevents scalability to many data sources and high data volume. We therefore need better support for incremental KG updates, especially in toolsets. Such a capability has to provide solutions to a variety of issues. As already discussed in Sec.~\ref{subsubsec:DataAqc}, it has to be detected \textit{if} there are changes in the input data and if so to determine \textit{what} has changed. 
The changes to be dealt with are not limited to the addition of new information but deletions and updates in the sources have to be propagated to the KG as well.  
Changes that impact the underlying ontology or the pipeline's configuration also have to be managed and may require manual interaction/confirmation.
Support for a streaming-like propagation of source changes should also increase so that KGs can provide the most recent information. 

\textbf{Lack of open tools} As we have seen in Table~\ref{tab:benchmarkDimensions} most KG construction toolsets, especially the more advanced ones, are closed source and can thus not be used by others for creating a KG or for evaluating their functionality. Hence there is a strong need for more open-source toolsets to help improve the development of KGs and to advance the state-of-the-art in KG construction. 
Researchers and developers providing such an implementation and associated publications could achieve a high impact~\cite{OpenSourceScienceImpact}.

\textbf{Improved Extensibility \& Modularisation, Ease of Use.} A toolset for KG construction should be able to define and execute different pipelines depending on the data sources to be integrated and specific requirements, e.g., for incremental updates. 
Hence, an extensible and modular KG construction approach should be provided with alternate implementations for the different KG construction tasks to choose from. 
This can be facilitated by making use of existing implementations as has been done already for NLP tasks (e.g., Stanford CoreNLP) but not yet for other tasks such as entity resolution. 
From the projects compared in Section~\ref{sec:example-kgs} only a few addressed this problem so that more solutions are needed. 

The definition of a KG construction pipeline should be relatively easy, supported by a user-friendly GUI and with a low effort for configuring the pipeline and its individual tasks. 
The configuration can be simplified by providing default settings for most parameters or even automatic approaches based on machine learning~\cite{mahdavi2019towards}. On the other hand, a manual configuration should also be possible to achieve customizability and support for special cases (e.g., a new entity type or input format).  
Extensibility and modularisation of a tool should not lead to a higher configuration effort for users.

\textbf{Data and metadata management.} Good data and metadata management is vital in an open and incremental KG process. 
Only a few solutions even mention an underlying management architecture supporting the construction processes. 
Having uniform access or interfaces to data and relevant metadata can drastically improve the quality of the former~\cite{wilkinson2016fair}
and increases the workflow's replicability and possibilities for debugging. 
A dedicated metadata repository can store used mappings, schemata, and quality reports, improving the transparency of the entire pipeline process.

Metadata support is limited in current solutions and only some pipeline approaches acknowledge the importance of provenance tracking and debugging possibilities. 
We found that the term provenance is rather vaguely used, mostly in the meaning of tracking the source of facts and the processes involved in their generation.
Only few approaches such as SAGA~\cite{ilyas2022saga} also try to maintain the trustworthiness of facts.
Metadata such as fact-level provenance should be used more to support construction tasks, such as for data fusion to determine final entity values.
In general there is a need for maintaining more metadata, especially temporal information, that is also essential for studying the evolution of KG information.
Support for developing temporal KGs maintaining historical and current data, compared to the common sequences of static KG snapshot versions, is also a promising direction. 

\textbf{Data Quality} One of the main goals of KG construction is to achieve and maintain a high quality KG. The difficulty of this task grows 
with rising number and heterogeneity of data sources, especially if one relies on automatic data acquisition and data integration. 
High-quality sources can provide a clean type hierarchy and can serve as training data to alleviate some data-quality issues, that would be more difficult to address by treating low-quality sources in isolation~\cite{Weikum2021MachineKC}. 
Lower quality data sources often contain a high degree of long tail entities (which is the reason these data source are valuable).
Nevertheless, corroborating the information from these sources with evidence from higher-quality sources remains difficult and can reduce data quality.
For example, the automatic fusion of conflicting entity values can easily introduce wrong information into a KG and even a restricted degree of human intervention is problematic on a large scale~\cite{osman2021ontology,DBLP:journals/www/ZhaoJLJS20KGFusion}.

To achieve the best possible data quality, data cleaning should be part of all major steps in the construction pipeline so that the degree of dirty or wrong information that is entering the KG is limited. Moreover, the identification and repair of errors should be a continuous task, especially in large KG projects~\cite{Shenoy2022WikidataQuality}.
To better address these problems, more comprehensive data quality measures and repair strategies are needed that minimize human intervention to retain high scalability for KG construction. 

\textbf{Evaluation} The evaluation of complete KG construction pipelines is an open but important problem to measure the performance and quality of current approaches and to improve on them. So far there are benchmarks for individual tasks such as knowledge extraction~\cite{Nuzzolese2015OpenKE,Rodrguez2016PerformanceEO,Zhang2017PositionawareAA}, ontology matching~\cite{Euzenat2011OntologyAE}, entity resolution~\cite{Kopcke2010Benchmark, eagerkgcw2021, OpenEA,hertling2022gollum} and KG completion~\cite{ILPC2022, Codex, OGBLSC}.
While these benchmarks in some cases still leave gaps, e.g, regarding scalability~\cite{christophides2020bigdataER} or domain diversity~\cite{Portisch2022BackgroundKI}, they are already quite complex and indicate the high difficulty of defining a benchmark for the entire KG construction pipelines. 

A benchmark could be based on similar settings than for the creation of specific KGs discussed in Section \ref{sec:example-kgs} aiming at the initial construction and incremental update of either a domain-specific or cross-domain KG from a defined set of data sources of different kinds. The KG ontology and the KG data model (RDF or property graph) could be predefined to facilitate the evaluation of the resulting KG. The size of the resulting KG should be relatively high and the construction should be challenging with the need of knowledge extraction, entity linking/resolution and entity fusion. Determining the quality of the constructed KG is difficult as it would ideally be based on a near-perfect result (gold standard) for the initial KG and for its updated version(s). For all entity and relation types in the given KG ontology, it has then to be determined to what degree they could correctly be populated compared to the gold standard which requires an extension to known metrics such as precision and recall. Further evaluation criteria include the runtimes for the initial KG construction and for the incremental updates and perhaps the manual effort to set up the construction pipelines. Ideally, an evaluation platform could be established - similar to other areas~\cite{HOBBIT} - for a comparative evaluation of different pipelines with different implementations of the individual construction steps.

\textbf{The whole is more than the sum of it's parts.} While the individual parts of KG construction pipelines are well-established research problems with sometimes decades of previous research, the complex interaction of the pipeline tasks is not well researched yet. For example, the disambiguation strategies of the knowledge extraction task, especially entity linking, are very similar to entity resolution. The use of background knowledge and various inter-dependencies between different information is commonly summarized as \textit{holistic entity linking}. This approach has seen some research attention and a survey with future research directions was published by Oliveira et. al. in a 2021 paper~\cite{OLIVEIRA2021101624}. While such approaches go in the right direction, our pipeline scenario would invite an even more holistic case, where named-entity linking and entity resolution approaches aid each other to boost their performance. Furthermore, data cleaning can independently be done in several tasks but it would be beneficial to have a coordinated approach to avoid duplicate efforts. 

\section{Related Work}
\label{sec:rel-work}
The construction of KGs uses technologies from different areas and we have  discussed the tasks and surveys in these areas already in Section~\ref{sec:tasks}. Here we therefore focus on related surveys on the construction of KGs in general.

On almost 300 pages, Weikum et. al.~\cite{Weikum2021MachineKC} give an extensive tour on the automatic creation and curation of RDF-based knowledge bases or KGs, specifically from semi- and unstructured sources.
Their discussion of requirements is also concerned on the KG itself, whereas our requirements are more focused on the KG construction process. We also cover structured input data for KG construction, e.g., in  the requirements on \textit{Input Data} and tasks such as entity resolution. 
Their article provides overviews about the open knowledge graphs YAGO, DBpedia, NELL, and Wikidata, which are also discussed in our work, as well as on industrial knowledge graphs, which we only mention briefly due to the limited amount of publicly available information. By contrast, we systematically compare many further approaches w.r.t our derived requirements including general KG construction toolsets. Furthermore, we have identified several new challenges for future work, e.g., regarding incremental approaches,  open toolsets and benchmarks. 

Hogan et. al.\cite{hogan2021knowledge} give a comprehensive introduction to KGs.
Similar to us, their discussion includes multiple graph data models, and they present methods to deal with unstructured, semi- and structured data.
Serving as an introductory text to KGs in general, they provide a broad view on KGs including tasks like learning on KGs or publishing them. We are more focused on KG construction and cover many additional aspects such as requirements for KG construction and maintenance, a more detailed discussion of construction tasks,  a systematic comparison of state-of-the-art approaches as well as open challenges for KG construction. 

Ryen et. al~\cite{ryen2022building} provide a systematic literature review on KG creation approaches based on Semantic Web technologies. They survey and compare  36 approaches w.r.t their identified construction steps: ontology development, data preprocessing, data integration, quality and refinement, and data publication. One of their findings was that data quality appears to be a major blind spot.
We more comprehensively investigate the requirements, approaches and open challenges of  KG construction and maintenance and also include other KG data models such as the PGM.

Tama\v{s}auskait\.{e} et. al.~\cite{Tamaauskait2022DefiningAK} propose a KG development life-cycle consisting of six main steps with several possible subtasks.
Our work covers their construction tasks and feasible solutions in more detail and complements them with other relevant main tasks such as metadata management and the discussion of temporal aspects and versioning. Furthermore, our survey and evaluation of KG construction approaches are based on a set of requirements that led to the tasks not covered in their approach. We also provide a comparison of many construction approaches and toolsets and identify open challenges not covered in their work.  

There are several papers that discuss the construction of KGs for specific cases.
Zhu et al.~\cite{Zhu2022MultiModalKG} focus on the creation of multi-modal KGs, specifically combining the symbolic knowledge in a KG with corresponding images. They comprehensively present the two directions in which this task can be performed: visual knowledge extraction in order to label images with information from the KG and discovering images that describe entities and relations from the KG.
In our work, we aim to more broadly discuss the KG construction process in a complementary manner and only briefly discuss multi-modal (image-related) techniques. 
Xiaogang Ma reviews applications and construction approaches for KGs in the geoscience domain~\cite{Ma2021KnowledgeGC}. The discussed KG creation methods range from mostly manual approaches to processes relying on data mining of crowdsourced data. Furthermore, they discuss how KGs are used in geoscience data analysis, e.g., to enhance information extraction for public health hazards.
\c{S}im\c{s}ek et. al~\cite{simsek2021knowledge} give a high-level overview of the KG construction process in the general context of a KG's lifecycle. Their discussion of these general steps is alongside an in-use case study, where they provide the challenges they encountered. While they give valuable insights into KG construction in the real world they do not include a systematic comparison of the state-of-the-art approaches w.r.t the requirements of KG construction.

In summary, our work focuses more on KG construction  than previous KG surveys and provides additional information in several areas related to KG construction. 
We are not limiting ourselves to RDF-based KGs but also consider alternate graph data models such as the PGM, we consider not only the acquisition  and integration of unstructured and semistructured data but also of structured data, and we 
not only consider the one-time construction of KGs but also their incremental maintenance. In contrast to most previous surveys, 
we explicitly specify the main requirements for KG construction and use these  as a guideline for evaluating  and comparing many  KG-specific construction approaches and toolsets and identifying new open challenges.

\section{Conclusion}

This work presented the current state of knowledge graph construction, giving an overview of the requirements and defining this area's central concepts and tasks.
We gave a synopsis of techniques used to address individual steps of such a pipeline with a perspective on how well the state-of-the-art solutions for these specific tasks can be integrated into an incremental KG construction approach. 
We comparatively analyzed a selection of current KG-specific pipelines and toolsets for KG construction, based on a list of criteria derived from our initial requirements set.
We found vast differences across these pipelines concerning the number and structure of the input data, applied construction methods, ontology management, the ability to continuously integrate new information and the tracking of provenance throughout the pipeline. 
The open KG-specific approaches are currently rather limited in their scalability to many sources, support for incremental updates and in several steps regarding metadata, ontology management, entity resolution/fusion, and quality assurance. The considered toolsets are generally better in terms of their functionality but they are mostly closed-source and thus not usable for new KG projects or research investigations.  

We identified various challenges that need to be addressed for improved incremental KG construction. 
These problems range from engineering questions like the need for a flexible software architecture, over numerous task-specific problems and the support for incremental construction, to hurdles that must be addressed collectively by the research community, like the development of open-source and modular toolsets for KG construction and on benchmarking and evaluation processes.
Concerning the exploitation of new data sources, integrating more multimodal data is of great potential but also requires more research to achieve effective solutions. 

Addressing the derived challenges promises significant advances for future KG construction pipelines and much reduced effort for creating and maintaining high-quality KGs.

\vspace{10mm}
\noindent
\textbf{Acknowledgements.}
The authors acknowledge the financial support by the Federal Ministry of Education and Research of Germany 
and by the S\"achsische Staatsministerium f\"ur Wissenschaft Kultur und Tourismus in the program Center of Excellence for AI-research 
"Center for Scalable Data Analytics and Artificial Intelligence Dresden/Leipzig", 
project identification number: ScaDS.AI 

\bibliographystyle{ios1}
\bibliography{cleaned_library}

\end{document}